\documentclass[pdflatex,sn-nature]{sn-jnl}

\usepackage{graphicx}%
\usepackage{multirow}%
\usepackage{amsmath,amssymb,amsfonts}%
\usepackage{amsthm}%
\usepackage{mathrsfs}%
\usepackage[title]{appendix}%
\usepackage{xcolor}%
\usepackage{textcomp}%
\usepackage{manyfoot}%
\usepackage{booktabs}%
\usepackage{algorithm}%
\usepackage{algorithmicx}%
\usepackage{algpseudocode}%
\usepackage{listings}%
\usepackage{tabularx}
\usepackage{array}
\usepackage{subcaption}
\usepackage{multirow}
\usepackage{tikz}
\usepackage{pgfplots}
\usepackage{helvet} 
\usepackage[final]{changes}
\usepgfplotslibrary{groupplots}
\pgfplotsset{compat=1.18}

\usepackage{bibunits}
\defaultbibliography{references}
\defaultbibliographystyle{sn-nature}

\usepackage{pifont}
\newcommand{\cmark}{\ding{51}} 

\definechangesauthor[name={Me}, color=blue]{me}
\setdeletedmarkup{\textcolor{red}{\sout{#1}}}

\definecolor{delayfixed}{HTML}{C3B4DC}    
\definecolor{delaylearned}{HTML}{5B2C8D}  
\definecolor{axisgray}{HTML}{4D4D4D}

\newcolumntype{C}{>{\centering\arraybackslash}X}

\theoremstyle{thmstyleone}%
\theoremstyle{thmstyletwo}%

\theoremstyle{thmstylethree}%

\algtext*{EndFor}
\algtext*{EndIf}
\algnewcommand\algorithmicinput{\textbf{Input:}}
\algnewcommand\Input{\item[\algorithmicinput]}
\algnewcommand\algorithmicparam{\textbf{Parameters:}}
\algnewcommand\Param{\item[\algorithmicparam]}
\algnewcommand\algorithmicoutput{\textbf{Output:}}
\algnewcommand\Output{\item[\algorithmicoutput]}

\begin{document}

\title[Article Title]{DelRec: learning delays in recurrent spiking neural networks}

\author*[1, 2, 3]{\fnm{Alexandre} \sur{Queant}}\email{alexandre.queant@cnrs.fr}

\author[1, 2]{\fnm{Ulysse} \sur{Rançon}}\email{ulysse.rancon@cnrs.fr}

\author[1, 2, 3]{\fnm{Benoit} \sur{R Cottereau}}\email{benoit.cottereau@cnrs.fr}
\equalcont{These authors contributed equally to this work.}

\author[1, 2]{\fnm{Timoth{\'e}e} \sur{Masquelier}}\email{timothee.masquelier@cnrs.fr}
\equalcont{These authors contributed equally to this work.}

\affil[1]{\orgdiv{CerCo}, \orgname{CNRS UMR 5549}, \city{Toulouse}, \country{France}}
\affil[2]{\orgdiv{Université de Toulouse}, \country{France}}
\affil[3]{\orgdiv{IPAL}, \orgname{CNRS IRL 2955}, \country{Singapore}}

\abstract{
Biological neurons transmit information with stereotyped electrical impulses called ``spikes'', sensitive to coincident timings. Spiking Neural Networks (SNNs), introduced in the nineties, have gained popularity in AI for their energy efficiency and competitive performance with deep learning. Among them, Recurrent SNNs (RSNNs) are particularly appealing for their ability to learn long-term dependencies and exhibit rich dynamics. In SNNs, each connection can have a weight and a transmission delay, both plastic in the brain. While theory has long suggested that trainable delays enhance a network's expressivity, practical learning methods emerged only recently and remain mostly limited to feedforward delays. Here, we introduce DelRec, the first method to jointly optimize recurrent delays with synaptic weights in RSNNs via surrogate gradient learning, compatible with any spiking neuron model. DelRec works in discrete time, leveraging differentiable interpolation to handle non-integer delays with well-defined gradients at training time. Using simple neurons, DelRec outperforms all baselines on a chaotic time-series prediction task, and achieves competitive performance on four challenging temporal datasets. Analysis of trained networks show that recurrent delay optimization builds structured, depth-dependent spatio-temporal receptive fields that the network actively exploits, with delay-weight joint optimization reshaping temporal selectivity. Beyond accuracy, we show that recurrent delay learning can reduce the memory footprint and energy consumption of a network, and shapes its robustness to temporal perturbations of its inputs. This work establishes recurrent delay optimization as a promising framework for both biological circuit modeling and neuromorphic computing.
}

\maketitle

\begin{bibunit}

\section{Introduction}\label{sec1}

\begin{figure}
  \centering
  \includegraphics[width=1\linewidth]{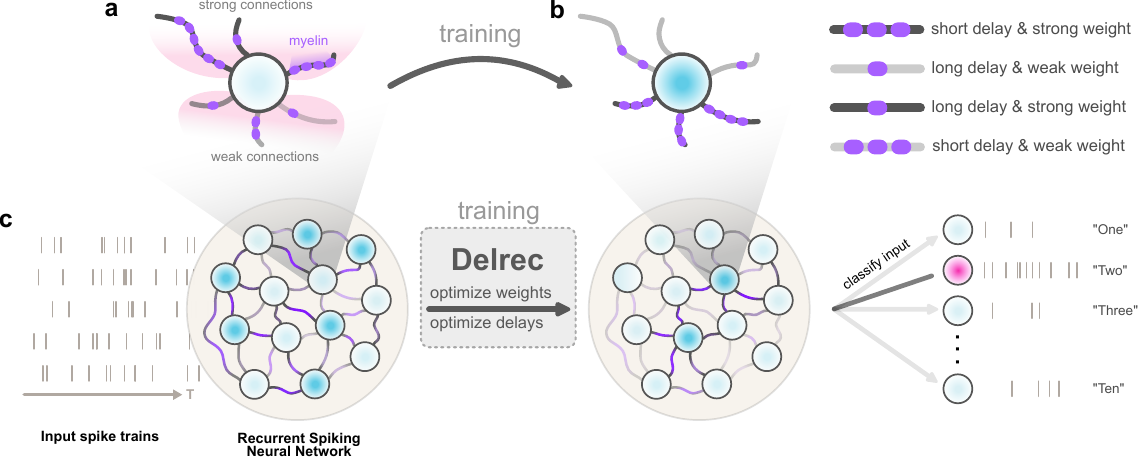}
  \caption{\textbf{Learning delays in recurrent connections as a mechanism for temporal specialization in RSNNs.} \textbf{a}, Biological motivation: activity-dependent myelination shapes the conduction delay between neurons, while activity-dependent synaptic plasticity shapes the connection strength. \textbf{b}, Delay-weight specialization with the joint optimization of both parameters in order to solve a task. Recurrent delays and synaptic weights allow the network to control both the latency and the magnitude of recurrent interactions. \textbf{c}, Overview of DelRec. Input spike trains (e.g., from temporal classification tasks) are processed by an RSNN, in which recurrent weights and delays are learned jointly to predict an output (e.g., the stimulus class). Node color indicates neural activity, whereas branch opacity and purple saturation denote synaptic strength and effective myelination, respectively. The evolution of weights and delays in the figure are illustrative, and do not result from any real optimization procedure}
  \label{fig:intro}
\end{figure}

The brain operates across a wide range of timescales: rapid feedforward processing enables the near-instantaneous recognition of sensory patterns, while sustained neural activity over longer periods supports higher-order cognitive functions such as working memory, decision-making and continuous motor control. This capacity for temporal integration is widely attributed to recurrent connections between neurons, which enable transient inputs to generate persistent or oscillatory activity that extends well beyond the duration of the stimulus \citep{Wang2001}. In particular, the timing of these recurrent interactions is governed by conduction delays between neurons, which are not simple anatomical constraints but potential degrees of freedom for structuring circuit dynamics. Axonal myelination, the main factor of conduction velocity, is itself plastic and experience-dependent \citep{Monje2018}, suggesting that neural transmission delays are continuously optimized through development and learning (see illustration in Fig~\ref{fig:intro}a,b). Consistent with this view, heterogeneous transmission delays within recurrent circuits are believed to play a central role in neural computation by shaping the temporal dynamics of recurrent networks and enhancing their computational capacity. For example, neural oscillations, whose phase relationships coordinate communication between neuronal populations, critically depend on the precise timing of recurrent interactions \citep{davisSpontaneousTravelingWaves2021}. Similarly, polychronization, the emergence of time-locked firing patterns among groups of neurons with specific delay configurations, requires specific heterogeneous delays and expands a network's representational capacity beyond what connectivity alone can provide, offering a substrate for memory and temporal coding \citep{izhikevichPolychronizationComputationSpikes2006}. Understanding how delays shape recurrent dynamics is therefore both a fundamental question in neurobiology and a challenge for neuromorphic computational models, where learning to exploit temporal structure could lead to more efficient and expressive architectures for processing temporal data.

Recurrent Spiking Neural Networks (RSNNs) provide a natural framework for exploiting these principles. Inspired by the architecture and dynamics of biological neural circuits, they are inherently designed to process time-varying signals. Their recurrent connectivity allows transient inputs to influence future network activity by maintaining an internal state, enabling temporal integration over extended timescales. This capability is essential for tasks with long-range temporal dependencies, including speech recognition or time-series prediction \citep{bellec}. However, despite their promise, RSNNs remain underutilized in machine learning due to significant training challenges, particularly the pervasive issues of vanishing and exploding gradients. Recent advances have sought to overcome these limitations by enriching spiking neuron models \citep{yin2021accurate, Bittar2022, Baronig2025}. In particular, adaptive leaky integrate-and-fire (AdLIF) neurons and other sophisticated neuronal dynamics have substantially increased the computational power of RSNNs, yielding state-of-the-art performance across a wide range of spiking neural network benchmarks. Yet, these neuron-centric approaches emphasize modulating current inputs rather than reactivating past signals, which limits their ability to model extended temporal dynamics.

The biological relevance of conduction delays and their role in neural dynamics motivate an alternative modeling strategy that introduces temporal structure directly into neural interactions. Indeed, theoretical work demonstrated that delays enhance network expressivity \citep{maassComplexityLearningSpiking1999}: neurons detect coincident spike-time latencies, which, in the presence of heterogeneous delays, correspond to arbitrary spike-onset latency sequences. While previous studies have successfully demonstrated the benefits of using delays in feedforward connections \citep{shrestha2018slayerspikelayererror, Sun2022a, Grappolini_2023, Sun2023, Timcheck2023, Hammouamri2024, deckers2024colearning, goltz2025delgrad, Ghosh2025, sun2025exploitingheterogeneousdelaysefficient}, their potential in recurrent connections remains largely unexplored. Recurrent delays could provide additional computational benefits by shaping self-sustained activity, extending memory over long timescales, and enabling richer internally generated temporal dynamics \citep{singer_rec_dyn}. Theoretical works \citep{gilsinn2009dde,izhikevichPolychronizationComputationSpikes2006} show that recurrent delays transform a neuron’s differential equation, expanding the range of possible solutions and enabling richer dynamics. Additionally, recurrent delays may mitigate gradient challenges by implementing temporal skip connections, improving gradient propagation during training.

To date, however, only a handful of studies have explored the potential of delays in recurrent connections, and even fewer have focused on learning optimal recurrent delay configurations. An obstacle to this extension lies in how Spiking Neural Networks simulators usually process time-series (see Supplementary~\ref{supp:Updating recurrent delays in SNNs simulators} for details). A few recent approaches have introduced algorithms to optimize these temporal parameters, demonstrating promising improvements in temporal tasks. For instance, \citep{xuASRCSNNAdaptiveSkip2025} achieved state-of-the-art results by learning a single recurrent delay parameter per layer using backpropagation. Their approach selects delays from a fixed set via a softmax function with a decreasing temperature, showcasing the potential of more flexible, parameterized methods. To the best of our knowledge, only \citep{meszaros2025efficienteventbaseddelaylearning} has proposed an algorithm specifically designed to learn optimal delays in recurrent connections. Their method, tailored for the EventProp algorithm \citep{Wunderlich2021Jun}, leverages exact gradient computation. However, it inherits common unresolved limitations of EventProp, including scalability challenges and suboptimal performance on real-world temporal benchmarks. Currently, all state-of-the-art spiking approaches on these benchmarks rely on surrogate gradient learning (SGL) \citep{Neftci2019a}. 

In this paper, we introduce ``DelRec'', the first method to train axonal or synaptic delays in recurrent connections using surrogate gradient learning (SGL) and backpropagation\footnote{Our code is available at \url{https://github.com/alexmaxad/DelRec}.}. The method operates in discrete time and eliminates the need to predefine a maximum delay range. During training, we relax the constraint of integer delays by employing a differentiable interpolation process (Section~\ref{sec:learning-delays}). Our approach is implemented using the Pytorch-based Spikingjelly library \citep{spikingjelly}, and is compatible with any spiking neuron model. Our contribution is threefold. First, we show that DelRec can achieve state-of-the-art performance on different temporal benchmarks using an equivalent or lower number of parameters than previous competitive approaches (Section~\ref{sec:tasks}), and simple Leaky-Integrate-and-Fire neurons. Then, we investigate how delays in recurrent connections and their optimization can modify information encoding and spatio-temporal selectivity in a network (Section~\ref{self orga}). Finally, we demonstrate that learning recurrent delays enables more efficient networks by reducing memory footprint and energy consumption while maintaining performance, and reveals how temporal structure can improve robustness to data perturbations (Section~\ref{sec:constraints}). DelRec thus establishes recurrent delays as a valuable computational resource for both neuromorphic engineering and computational neuroscience. In neuromorphic systems, our results highlight the potential of exploiting delays in recurrent spiking neural networks for efficient temporal processing. Using DelRec, these delays can be optimized off-chip effectively and implemented on chips with programmable delays, an approach whose efficiency has already been demonstrated by \citep{meszaros2025completepipelinedeployingsnns}. For biological modeling, it provides a computational framework for studying how adaptive delays interact with synaptic strengths to shape recurrent dynamics across timescales. Taken together, our results suggest that the timing of recurrent interactions is a valuable parameter to optimize, both for building sparse and frugal temporal spiking networks and as an important mechanism of adaptation in the brain.

\section{Results}

\subsection{Learning delays in recurrent spiking neural networks}\label{sec:learning-delays}

\noindent\textbf{Model.} Most spiking neuron models can be described by three discrete-time equations \citep{fang2021incorporating}:
\begin{align}
H[t] &= f(V[t-1], I[t]), \label{eq neuronal charge} \\
S[t] &= \Theta(H[t] - V_\mathrm{th}), \label{eq neuronal fire} \\
V[t] &= \begin{cases}
H[t]\cdot (1 - S[t]) + V_\mathrm{reset}\cdot S[t], \mathrm{~~if~hard~reset}\\
H[t] - V_\mathrm{th}\cdot S[t],\mathrm{~~if~soft~reset} \\
\end{cases}, \label{eq neuronal reset}
\end{align}
where $f$ is the neuronal charge function, $I[t]$ is the input current, $H[t]$ is the membrane potential after charging but before firing, $V[t]$ is the membrane potential after firing, and $S[t]$ is the output spike. $V_\mathrm{th}$ is the firing threshold and $V_\mathrm{reset}$ is the reset potential. $\Theta(x)$ is the Heaviside step function. We use the surrogate gradient method \citep{Neftci2019a} to handle its non-differentiability during training. In all our experiments, we use the leaky integrate-and-fire (LIF) neuron, whose charge function is $f(V[t-1],I[t]) = (1- \frac{1}{\tau}) \cdot V[t - 1] + \frac{1}{\tau} \cdot I[t]$, although our method is compatible with any neuron model fitting the above formalism.

The total input current $I[t] = X[t] + X^\text{rec}[t]$ is the sum of feedforward inputs $X$ and the recurrent component $X^\text{rec}$. In a standard Recurrent Spiking Neural Network (RSNN), the recurrent input to neuron $i$ in a layer $L$ of neurons is:
\begin{equation}\label{eq:2}
X_i^{\text{rec}}[t] = \sum_{j \in (L)} w_{ij}^{\text{rec}} S_j[t-1]
\end{equation}
where $\{w_{\text{rec}}\}$ are the recurrent weights and $(L)$ denotes the set of indices of neurons belonging to layer $L$. We extend this formulation by introducing learnable delay parameters $\{d_j \in \mathbb{N} , \, j\in L \}$: a spike emitted by neuron $j$ at time $t$ now reaches neuron $i$ at time $t+1+d_j$, yielding:
\begin{equation}\label{eq:3}
X_i^{\text{rec}}[t] = \sum_{j \in (L)} w_{ij}^{\text{rec}} S_j[t-(1+d_j)].
\end{equation}
Eq.~\ref{eq:3} assigns a single delay $d_j$ to all outgoing connections of neuron $j$, which we call an axonal delay, by analogy with a conduction delay set by the myelination of a single axon. Our method also handles synaptic delays, one per connection, obtained by replacing $d_j$ with $d_{ij}$. Both are learned by the same procedure and differ in granularity and cost: a layer of $N$ neurons carries $N$ axonal delays but $N^2$ synaptic ones. A delay parameter $d = 0$ corresponds to the standard RSNN case with an effective delay of one time step. For clarity, we write the remainder of this section for axonal delays, the synaptic case follows by replacing $d_j$ with $d_{ij}$ throughout, and both configurations are reported in all our experiments.

The introduction of delays $d_j$ in Eq.~\ref{eq:3} places the recurrent dynamics of a spiking layer with delayed connections within the framework of delay differential equations (DDEs), in which the state of the approximated, time-continuous system depends not only on its present but also on its past values. In this setting, in this approximated system, the introduction of a delay transforms a finite-dimensional dynamical system into an infinite-dimensional one: the system's state is no longer a point, but a function over a continuous interval of past values \citep{gilsinn2009dde}. This profoundly enriches the accessible dynamics: a simple scalar delayed differential equation can spontaneously generate sustained periodic oscillations that are entirely absent from its non-delayed counterpart, where the system can only converge to a fixed point. When combined with nonlinearities, as is inherently the case with the thresholding and reset mechanisms in Eq.~\ref{eq neuronal fire}-\ref{eq neuronal reset}, delays further enable the coexistence of multiple stable dynamical regimes \citep{gilsinn2009dde}.

These mathematical properties translate directly into enhanced computational capacity. Ref.\citep{izhikevichPolychronizationComputationSpikes2006} demonstrated that in a recurrent spiking network with heterogeneous conduction delays, specific subsets of neurons develop reproducible time-locked firing sequences called polychronous groups, where the precise delay structure causes spikes to arrive coincidentally at downstream neurons. The number of such polychronous groups far exceeds the number of neurons in the network, implying that the delay configuration itself acts as a vast reservoir of representational capacity.
\\

\noindent\textbf{Scheduling and learning procedure.} To leverage this expressivity, we introduce a method to jointly optimize the delay parameters $\{d_j\}$ with the recurrent weights $\{w_{ij}^{\text{rec}}\}$ using surrogate gradient learning and backpropagation. We adopt a future-oriented perspective: when neuron $j$ fires a spike at time $t$, we schedule a recurrent input $w_{ij}^{\text{rec}} S_j[t]$ at the future time $t + 1 + d_j$ for all neurons $i$ in the same layer. In practice, this is implemented through a scheduling buffer $X^{\text{rec}} \in \mathbb{R}^{N \times T}$ that stores weighted spikes for future delivery:
\begin{equation}
X^{\text{rec}}_{i}[t+\tau] = \sum_{ j \mid 1+d_j = \tau } w_{ij}^{\text{rec}} S_j[t] \label{eq:7}
\end{equation}

For optimization, we relax the delays to real values $\{d_j \in \mathbb{R}\}$. A spike scheduled at a non-integer time $t+1+d$ is split between the two neighboring integer time steps in proportion to their distance to $t+1+d$, so that the delay parameter receives a well-defined gradient: moving $d$ continuously transfers weight from one time step to the next. This linear interpolation is the required element to make recurrent delays trainable. 

\begin{figure}
  \centering
  \includegraphics[width=0.95\linewidth]{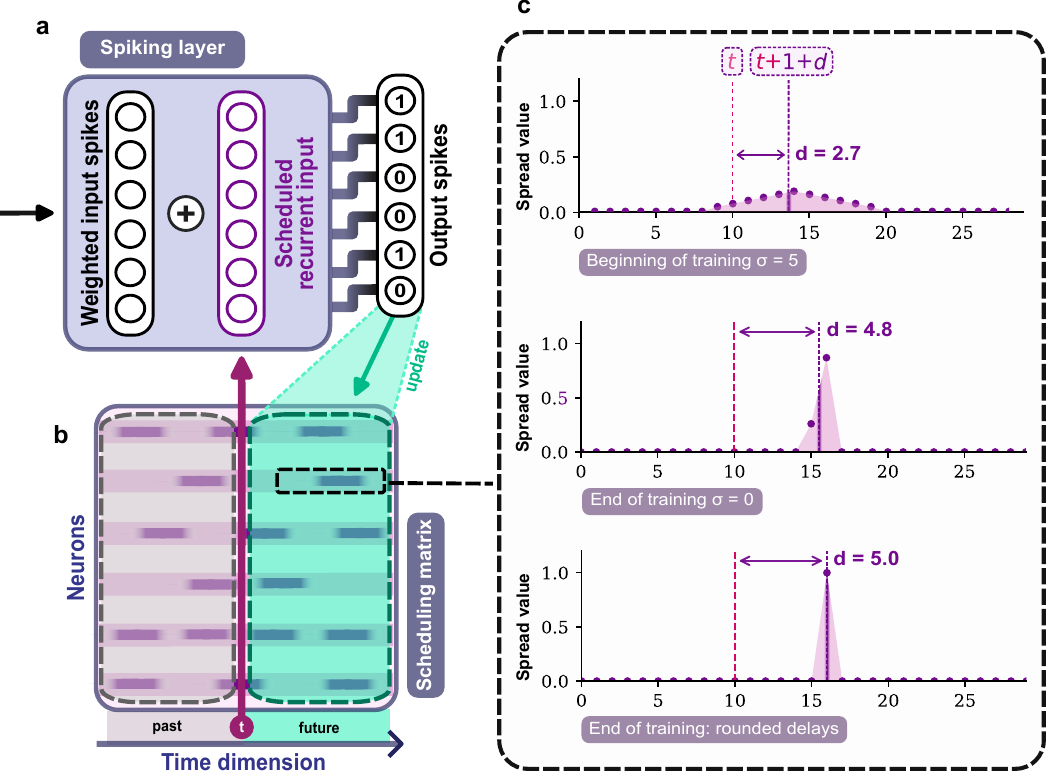}
  \caption{\textbf{Future spikes are weighted by the triangle function and stored in the scheduling matrix.} \textbf{a}, At time step $t$, the neurons in the studied layer receive weighted spikes from the previous layer, which are then summed with the inputs scheduled for time step $t$ in the scheduling matrix. The subsequent evolution of the internal state of the neuron (membrane potential) may produce output spikes. \textbf{b}, Each neuron of the layer receives a weighted sum of the output spikes, and schedules it on a spread of future dates determined by its delay parameter $d$ and the spread of the epoch $\sigma$. Spread values are represented by the purple gradients. \textbf{c}, We modify the spread function at each epoch by reducing its $\sigma$. At the beginning of training, the scheduled values are widely spread around the true delay $d$. When the training ends, $\sigma$ is close to $0$, and the spread function only performs linear interpolation between the closest integers from the floating delay. Then we manually round the delay to the nearest integer in evaluation mode.}
  \label{fig:scheduling matrix}
\end{figure}

To improve the smoothness of the scheduling, a spike scheduled at non-integer time $t + 1 + d$ can be temporally spread over more than two neighboring time steps using a triangle function $h_{\sigma, d}$ \citep{Khalfaoui2023_ICML} with width parameter $\sigma$
\begin{equation}
h_{\sigma, d}(\tau) = \max\!\left(0,\; \frac{1+\sigma - |\tau - (1+d)|}{(1+\sigma)^2}\right)
\label{eq:10}
\end{equation}
At each time step $t$, the scheduling matrix is updated for all future times $t + \tau$:
\begin{equation}\label{eq:rec_update}
X^{\text{rec}}_{i}[t+\tau] \leftarrow X^{\text{rec}}_{i}[t+\tau] + \sum_{j=1}^N w_{ij}^{\text{rec}} \cdot h_{\sigma_{\text{epoch}}, d_j}(\tau) \cdot S_j[t]
\end{equation}
The parameter $\sigma$ is progressively decreased throughout training (Fig.~\ref{fig:scheduling matrix}C), so that by the end of training the spread function reduces to a linear interpolation between the two closest integer positions. Because of the finite support of $h_{\sigma,d}$ (see Eq.~\ref{eq:supp_h}), scheduling only needs to cover a limited temporal range $\tilde{\mathbf{E}}(\sigma, D)$, whose size depends on the current maximum delay and the spread $\sigma$ (Eq.~\ref{eq : define-E-tilde}). We exploit this by implementing $X^{\text{rec}}$ as a circular buffer with a pointer mechanism (Algorithm~\ref{alg:learn-delays}). The initial broad spread allows the network to explore long-range temporal dependencies early in training, while the progressive narrowing refines delays to precise locations. This annealing strategy follows \cite{Hammouamri2024}. A detailed analysis of the computational overhead is provided in Supplementary~\ref{computational_overhead}. The complete mechanism is illustrated in Fig.~\ref{fig:scheduling matrix}.

While the simulation and the optimization of delays in recurrent connections introduce significant implementation challenges (Supplementary~\ref{supp:Updating recurrent delays in SNNs simulators}), they also offer a structural advantage for gradient-based training: in their presence, the computational graph is equipped with temporal skip-connections (Supplementary Fig.~\ref{fig:hard_rec_delays}b, bottom) bridging distant time steps without passing through all intermediate states. This allows shorter gradient paths between distant events, improving gradient flow during backpropagation through time (BPTT), which we explore in Section~\ref{self orga}.

\subsection{Recurrent delay learning improves temporal processing} \label{sec:tasks}

We evaluate DelRec on a controlled time-series task and on four classification benchmarks. The controlled task, forecasting the Mackey-Glass chaotic system \citep{harnessing_jaeger}, lets us modulate the temporal difficulty directly and observe how the benefit of delay learning depends on it. The four benchmarks then test whether the same effect holds on real-world tasks: the widely used Spiking Speech Commands (SSC) and permuted sequential MNIST (PS-MNIST) datasets, and two datasets introduced by \citep{chenNeuromorphicSequentialArena2025} to probe temporal
processing, Autonomous Localization (AL) and Human Activity Recognition (HAR). Complete dataset information, implementation details and hyperparameters are given in Supplementary~\ref{sec:datasets}. Throughout, we denote by Ff. and Rec. delays placed on feedforward and recurrent connections respectively, and distinguish axonal (ax., one delay per neuron) from synaptic (syn., one delay per connection) delays. Delays are learned jointly with the weights unless marked as fixed (fix.) at their
initialization. To isolate the contribution of delay learning itself, we compare  learned delays against fixed random delays in all experiments (Table~\ref{tab:ablation}), a suitable control that is often omitted in the delay learning literature.
\\

\begin{figure}
  \centering
  \includegraphics[width=\linewidth]{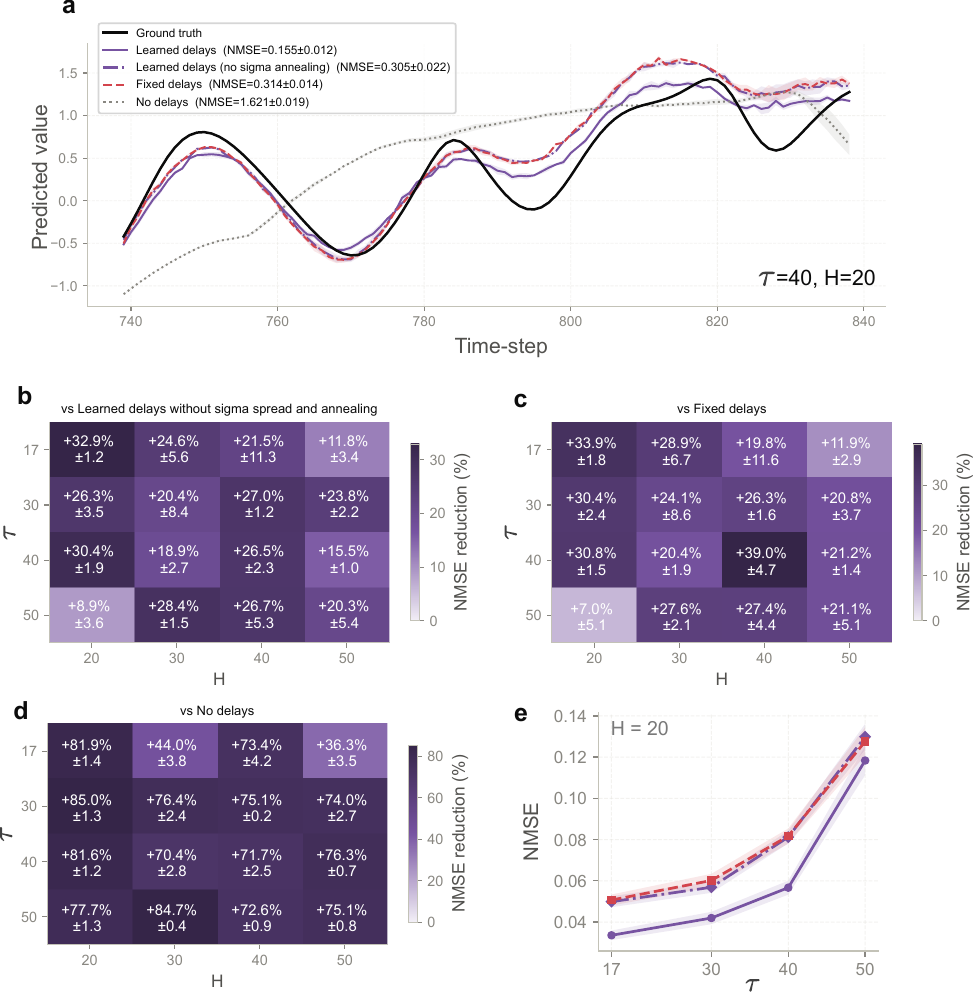}
  \caption{\textbf{Learned recurrent delays improve prediction on the
  Mackey-Glass chaotic system across all temporal difficulty conditions.}
  \textbf{a}, Predicted trajectories of the four variants against the ground truth at $\tau=40$, $H=20$ (mean over 3 seeds, shaded SEM, NMSE computed on the time-window).
  \textbf{b-d:} Relative NMSE reduction of the full DelRec model over three baselines, as a
  function of the chaos parameter $\tau$ and the prediction horizon $H$
  (mean $\pm$ SEM over 3 seeds). \textbf{b}, vs.\ a learned-delays model without $\sigma$-annealing, isolating the contribution of the annealing schedule. \textbf{c}, vs.\ fixed
  recurrent delays, isolating the contribution of delay learning.
  \textbf{d}, vs.\ an RSNN without delays, isolating the total contribution of recurrent delays. \textbf{e}, Absolute test NMSE as a function of $\tau$ at fixed horizon $H=20$, for the full model (learned delays), the same model without $\sigma$-annealing, and the fixed-delays baseline (mean over 3 seeds and SEM as shaded area).}
  \label{fig:mackey-glass}
\end{figure}

\noindent\textbf{Forecasting the Mackey-Glass chaotic system.} The Mackey-Glass system is governed by a delay differential equation,
\begin{equation}
\dot{x}(t) = \beta \frac{x(t-\tau)}{1 + x(t-\tau)^n} - \gamma x(t),
\label{eq:mackey-glass}
\end{equation}
whose delay parameter $\tau$ directly controls the dynamical complexity: for small $\tau$ the attractor is periodic, whereas for $\tau \gtrsim 17$ the dynamics become chaotic (Fig.~\ref{fig:mackey-glass}a). Increasing $\tau$ therefore extends the temporal dependencies a model must capture to predict future values, providing a tunable probe of delay learning in recurrent dynamics. We generate discrete time series by numerically integrating Eq.~\ref{eq:mackey-glass} and subsampling at unit intervals (see Methods~\ref{mg_methods} for all parameter values). Models receive windows of $150$ time-steps and predict the value at a horizon $H$, then we sweep over $\tau$ and $H$ values and compare four variants with a single hidden layer of 128 neurons: an RSNN without delays, an RSNN with fixed recurrent delays, an RSNN with learned recurrent delays but without the $\sigma$-annealing presented in Section~\ref{sec:learning-delays}, and the complete DelRec model (see Methods~\ref{mg_methods}). Unlike the classification tasks below, this benchmark requires continuously tracking a non-linear trajectory rather than collapsing temporal information into a single label, making it particularly suited to studying recurrent temporal dynamics.

Across all $(\tau, H)$ conditions, DelRec outperforms every baseline, reducing the NMSE over both a standard RSNN (Fig.~\ref{fig:mackey-glass}c) and a fixed delays model (Fig.~\ref{fig:mackey-glass}b). This gain relies significantly on the $\sigma$-annealing schedule (Fig.~\ref{fig:mackey-glass}a): at $H=20$, for example, the learned model tracks the fixed delay baseline, and only the annealed one improves on it (Fig.~\ref{fig:mackey-glass}d), validating the mechanism of Section~\ref{sec:learning-delays}. Fig.~\ref{fig:mackey-glass}e 
displays an example of the chaotic system and each model's prediction over that window. Full NMSE values across all $\tau$ and $H$ are given in Supplementary~\ref{sup:mg-bars} and corroborate these findings.
\\

\noindent\textbf{Comparison with the literature.} We evaluated our method on the four following benchmarks: the Spiking Speech Commands (SSC) and permuted sequential MNIST (PS-MNIST) datasets, which are widely used datasets in the SNN community, and both demand leveraging spatio-temporal patterns to reach good classification accuracies. In addition to these benchmarks, we further evaluate DelRec on two datasets introduced by \citep{chenNeuromorphicSequentialArena2025} that were specifically designed to probe temporal processing: the Autonomous Localization (AL) dataset and the Human Activity Recognition (HAR) dataset (see Supplementary~\ref{sec:datasets} for a details description). These two datasets target distinct temporal regimes: AL is a binary classification task that requires integrating information across an entire sequence of actions, thereby probing long-range temporal dependencies. Conversely, the HAR classes that are hardest to separate depend on short-timescale temporal structures embedded in noisy sensor recordings. Importantly, the benchmark of \citep{chenNeuromorphicSequentialArena2025} evaluates every model within a single shared pipeline and at a matched parameter budget of $\approx100$k, so that differences in accuracy reflect the temporal processing mechanism rather than model size or training setup, making it well-suited to a fair comparison of DelRec with LIF against other spiking neuron models and architectures. 

\begin{table}
\centering
\caption{\textbf{Comparison with the literature on SSC and PS-MNIST.}
Accuracies are mean $\pm$ std over the indicated number of seeds. A single
value denotes a single reported run. DelRec rows report our best delay
configuration, the full ablation is given in Table~\ref{tab:ablation}.}
\label{tab:SSC-PSMNIST}
\renewcommand{\arraystretch}{1.1}
\setlength{\tabcolsep}{4pt}
\footnotesize
\begin{tabularx}{\textwidth}{@{} X l c l r c r @{}}
\toprule
Model & Neuron & Rec. & Delays & Param & Seeds & Test Acc.\ [\%] \\
\midrule
\multicolumn{7}{@{}l}{\textbf{SSC}} \\
\addlinespace[0.15em]
Adaptive RSNN \citep{yin2021accurate}                          & AdLIF       & \cmark & ---         & 0.78M             & 1  & 74.20 \\
EventProp \citep{meszaros2025efficienteventbaseddelaylearning} & LIF         &        & Ff.         & $\sim$5M\tnote{a} & 8  & 76.1$\pm$1.0 \\
Bullet trains \citep{morrill2026bullet}                        & LIF         &        & Ff.         & n/r\tnote{b}      & 5  & 77.35$\pm$0.23 \\
RadLIF \citep{Bittar2022}                                      & RadLIF      & \cmark & ---         & 3.9M              & 1  & 77.40 \\
cAdLIF \citep{deckers2024colearning}                           & cAdLIF      &        & ---         & 0.35M             & 1  & 77.50 \\
d-cAdLIF \citep{deckers2024colearning}                         & cAdLIF      &        & Ff.         & 0.7M              & 10 & 80.23$\pm$0.07 \\
SE-adLIF \citep{Baronig2025}                                   & SE-adLIF    & \cmark & ---         & 1.6M              & 20 & 80.44$\pm$0.26 \\
DCLS \citep{Hammouamri2024}                                    & LIF         &        & Ff.\ (syn.) & 2.5M              & 5  & 80.69$\pm$0.21 \\
ASRC-SNN \citep{xuASRCSNNAdaptiveSkip2025}                     & LIF         & \cmark & Rec.(one per layer)        & 0.37M             & 1  & 81.54\tnote{c} \\
SiLIF \citep{fabreStructuredStateSpace2025}                    & SiLIF (SSM) & \cmark & ---         & 0.35M             & 10 & 82.03$\pm$0.25 \\
\textbf{DelRec} \textit{(ours)}                                & LIF         & \cmark & Rec.\ (ax.) & 0.37M             & 5  & \textbf{82.42$\pm$0.23} \\
\addlinespace[0.3em]
\midrule
\multicolumn{7}{@{}l}{\textbf{PS-MNIST}} \\
\addlinespace[0.15em]
GLIF \citep{yao2022glif}                                       & GLIF        & \cmark & ---         & 0.15M             & 1  & 90.47 \\
Adaptive RSNN \citep{yin2021accurate}                          & AdLIF       & \cmark & ---         & 0.15M             & 1  & 94.30 \\
BRF \citep{higuchi2024balancedresonateandfireneurons}          & RF          & \cmark & ---         & 69k               & 1  & 95.20 \\
ASRC-SNN \citep{xuASRCSNNAdaptiveSkip2025}                     & LIF         & \cmark & Rec.(one per layer)        & 0.15M             & 1  & 95.77\tnote{c} \\
\textbf{DelRec} \textit{(ours)}                                & LIF         & \cmark & Rec.\ (syn.) & 0.25M             & 5  & \textbf{96.63$\pm$0.12} \\
\bottomrule
\end{tabularx}
\begin{tablenotes}
\footnotesize
\item[a] Parameter count not stated in the original publication, estimated from their Figure 6.
\item[b] Not reported.
\item[c] Reproduced with the publicly available code, using dedicated validation and test sets.
\end{tablenotes}
\end{table}

\begin{table}
\centering
\caption{\textbf{Comparison with the literature on AL and HAR}. Baseline numbers are taken from \citep{chenNeuromorphicSequentialArena2025} (single runs, same splits and
protocol). All models use $\approx100$k parameters. Accuracies for DelRec are mean
$\pm$ std over 3 seeds for HAR and 5 for AL. The full ablation is given in Table~\ref{tab:ablation}.}
\label{tab:AL-HAR}
\renewcommand{\arraystretch}{1.1}
\setlength{\tabcolsep}{4pt}
\footnotesize
\begin{tabularx}{\textwidth}{@{} X l l c c @{}}
\toprule
 & & & \multicolumn{2}{c}{Test Acc.\ [\%]} \\
\cmidrule(l){4-5}
Model & Neuron & Delays & AL & HAR \\
\midrule
RSNN                                                                   & LIF       & ---          & 56.50 & 77.32 \\
Spike-Driven Transformer\tnote{a} \citep{yao2023spikedriventransformer} & LIF      & ---          & 58.00 & 71.07 \\
CE-LIF \citep{CE_LIF}                                                  & CE-LIF    & ---          & 60.62 & 80.83 \\
GSN \citep{hao2024ultralowpowerneuromorphicspeechenhancement}          & LIF       & ---          & 67.22 & 82.31 \\
Spiking TCN\tnote{a} \citep{bai2018empiricalevaluationgenericconvolutional}     & LIF       & ---          & 69.88 & 82.41 \\
LTC \citep{yin-LTC}                                                    & LTC       & ---          & 79.04 & 82.03 \\
Gated Spiking Unit \citep{stan2023learninglongsequencesspiking}        & LIF       & ---          & 80.42 & 89.16 \\
Binary S4D \citep{stan2023learninglongsequencesspiking}                & LIF (SSM) & ---          & 81.44 & \textbf{89.37} \\
\textbf{DelRec} \textit{(ours)}                               & LIF       & Rec.\ (ax.) & \textbf{85.74$\pm$1.49} & 83.53$\pm$0.26 \\
\bottomrule
\end{tabularx}
\begin{tablenotes}
\footnotesize
\item[a] The non-recurrent models in the table, all others use recurrent connections.
\end{tablenotes}
\end{table}

\begin{table}
\centering
\caption{\textbf{Learned and fixed recurrent delays on all four
benchmarks.} Within each dataset, all models share the same architecture,
hyperparameters, pre-processing, and seeds, and differ only in their delay
configuration: axonal or synaptic recurrent delays, either fixed at
initialization or learned jointly with the weights. Values are mean
$\pm$ std over 5 seeds, except for HAR, which is done over 3 seeds. The best configuration per dataset is in bold.}
\label{tab:ablation}
\renewcommand{\arraystretch}{1.1}
\setlength{\tabcolsep}{4pt}
\footnotesize
\begin{tabularx}{\textwidth}{@{} X l *{4}{c} @{}}
\toprule
 & & \multicolumn{4}{c}{Test Acc.\ [\%]}\\
\cmidrule(l){3-6}
Delay type & Learning & SSC & PS-MNIST & AL & HAR \\
\midrule
Axonal (ax.)    & Fixed   & 82.11$\pm$0.16          & 95.78$\pm$0.27          &     82.72$\pm$1.62      & 82.73$\pm$0.37 \\
Axonal (ax.)    & Learned & \textbf{82.42$\pm$0.23} & 96.32$\pm$0.19          &     \textbf{85.74$\pm$1.49}      & 83.53$\pm$0.26 \\
Synaptic (syn.) & Fixed   & 82.09$\pm$0.22         & 95.90$\pm$0.22          &     83.74$\pm$1.24    & 82.89$\pm$0.27\\
Synaptic (syn.) & Learned & 82.24$\pm$0.19          & \textbf{96.63$\pm$0.12} & 85.00$\pm$0.85 & \textbf{84.44$\pm$0.27} \\
\bottomrule
\end{tabularx}
\end{table}

On SSC and PS-MNIST, DelRec matches or exceeds the best published results while
using standard LIF neurons and a competitive parameter budget
(Table~\ref{tab:SSC-PSMNIST}). This result contrasts with many competing approaches that rely on more complex neuron models incorporating adaptive mechanisms, resonant dynamics, or structured state-space formulations \citep{Baronig2025, fabreStructuredStateSpace2025, higuchi2024balancedresonateandfireneurons}. This highlights the importance of incorporating delays in SNNs (see also \citep{Hammouamri2024, meszaros2025efficienteventbaseddelaylearning, xuASRCSNNAdaptiveSkip2025} for similar approaches), and suggests that even better results could be achieved by combining delays with more sophisticated neuron models. The same conclusion holds on the AL and HAR datasets: adding recurrent delays to an otherwise identical LIF network drastically raises the network's performance (from $56.5\%$ to $\approx82.7\%$ for AL and from $\approx77.3\%$ to $\approx82.7\%$ for HAR), while the delay learning algorithm further tunes the temporal structure and adds consistent and statistically significant improvement on top (Table~\ref{tab:ablation}, Supplementary~\ref{sup:significance}). DelRec remains below the two best HAR baselines \citep{stan2023learninglongsequencesspiking}, which replace the spiking neuron by a structured state-space backbone and, for the GSU, drop binary spiking in favor of ternary and continuous activations.

Our findings demonstrate that optimizing delays enhances the performance of spiking models and reveal that recurrent delays can provide greater computational benefits than feedforward delays, especially for tasks requiring long-range temporal integration. The relative benefit of the two delay types (axonal or synaptic), however, is not consistent across datasets. Synaptic delays, for example, which bring no benefit on the SSC or AL, outperform axonal ones on PS-MNIST and HAR tasks. This inconsistency, together with the large standard deviations of the learned-delays models on some datasets (e.g., $\pm1.49$ on AL) suggests that optimization becomes more challenging as the temporal structure of the model becomes increasingly parameterized. Whether this highlights an optimization difficulty or a genuine over-parametrization of the model's temporal structure remains an open question, and is a valuable direction for future work.

\subsection{Recurrent delay learning builds functional temporal structure}\label{self orga}

Behind the raw capacities of models on the benchmarks of Section~\ref{sec:tasks} lies an internal encoding that can differ with the architecture and the optimization algorithm used. Recurrent delays can act on this encoding in two distinct ways: their presence modifies the computational graph over which credit is assigned, while their optimization can be leveraged to change how information is encoded in the network. We study these two aspects in networks trained on the SSC dataset, focusing on the axonal models. 
\\

\begin{figure}
  \centering
  \includegraphics[width=\linewidth]{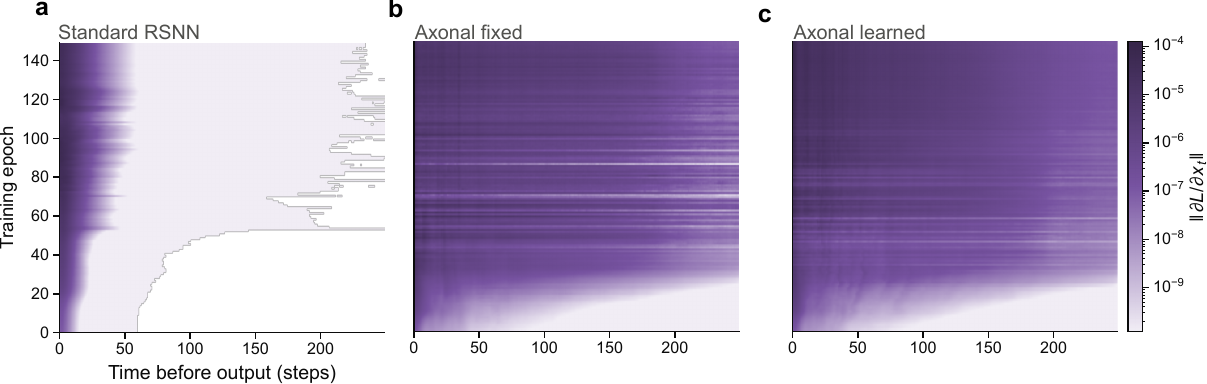}
  \caption{\textbf{Recurrent delays keep error gradients alive at long temporal depth throughout training.} Norm of the gradient of the final-step probe loss with respect to the input at a past time step, $|\partial \mathcal{L}_{\text{probe}}/\partial x_t|$, as a function of temporal depth (time before output, x-axis) and training epoch (y-axis) on SSC \textbf{a}, standard RSNN without delays. \textbf{b}, Fixed random axonal recurrent delays. \textbf{c}, Learned axonal recurrent delays. Panels \textbf{a–c} share a common logarithmic color scale. Values are averaged over a fixed batch of 245 test-set samples balanced across the 35 classes (7 per class).}
  \label{fig:gradflow}
\end{figure}

\noindent\textbf{Recurrent delays extend the temporal reach of credit assignment.} Adding delays to the recurrent connections modifies the computational graph by creating temporal skip-connections, as described in Section~\ref{sec:learning-delays}. We first probed the consequence of this modification on gradient propagation by logging, at every epoch and on a fixed class-balanced batch, the norm $|\partial \mathcal{L}_{\text{probe}}/\partial x_t|$ of the gradient reaching the input at each past time step (Fig.~\ref{fig:gradflow}). The probe loss is the cross-entropy of the readout at the final time step alone rather than the time-summed training loss, so that we can isolate the gradient flow that propagates through the recurrent dynamics (Methods~\ref{sec:methods_gradflow}). A standard RSNN loses the gradient signal beyond
$\sim$50 steps before the output at every epoch (Fig.~\ref{fig:gradflow}a), whereas recurrent delays (fixed or learned) keep meaningful gradients across the full temporal horizon (Fig.~\ref{fig:gradflow}b,c), showing that the presence of delays in recurrent connections suffices to propagate higher amplitude gradients further in time.
\\

\begin{figure}
  \centering
  \includegraphics[width=1\linewidth]{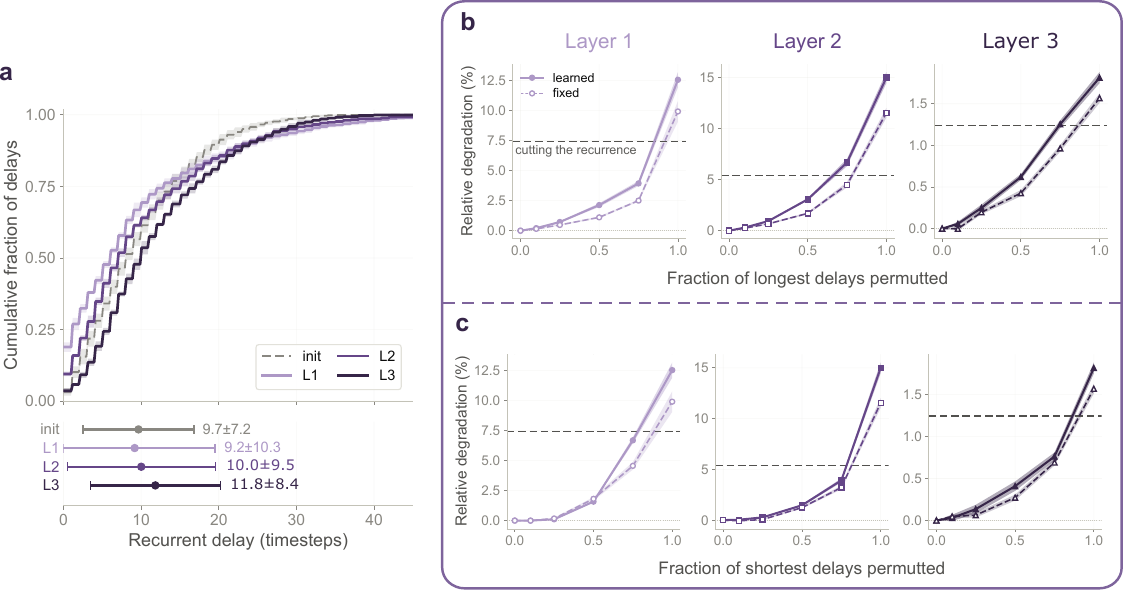}
  \caption{\textbf{Learned recurrent delays organize into a structure that the network exploits.} \textbf{a}, Cumulative distributions (top) and mean\,$\pm$\,std (bottom) of the learned axonal recurrent delays in each of the three hidden layers, together with the shared initialization (init, dashed). Shaded bands are std over 5 seeds, and delay distributions are cropped at the $99.5$th percentile. \textbf{b-c}, Permutation ablation: starting from a trained network, we select a fraction $q$ of each layer's neurons, either the $q$ with the longest delays (\textbf{b}) or the $q$ with the shortest (\textbf{c}), randomly permute their recurrent delays across those neurons (weights frozen), and measure the resulting relative accuracy degradation as a function of $q$, for the learned (solid) and fixed random (dashed) delays models. Curves and shaded bands are mean\,$\pm$\,SEM over the 5 seeds of trained models and 3 permutation draws, and the dashed horizontal line marks the degradation from removing the layer's recurrence entirely, averaged over the learned and fixed models.}
  \label{fig:delays_spe}
\end{figure}

\noindent\textbf{Delay optimization organizes delays into a structured configuration on which the network relies to perform temporal computations.} We investigated how delay optimization reshapes the recurrent temporal structure of the trained network, and whether the resulting structure is used. Fig.~\ref{fig:delays_spe}a shows that learning moves the recurrent delays away from their initialization in a depth-dependent way. The mean delay increases with depth, while the shallowest layer keeps the widest spread, with most neurons at short delays and a sparse tail reaching far longer ones. To test whether the network effectively relies on this structure, and on which delays, we permuted an increasing fraction $q$ of a layer's delays across neurons with the weights frozen, selecting either the $q$ longest delays first (Fig.~\ref{fig:delays_spe}b) or the $q$ shortest first (Fig.~\ref{fig:delays_spe}c). The permutation is done one layer at a time, leaving every other layer at its trained delays (Methods~\ref{sec:methods_permutation}). Because this preserves each layer's marginal delay distribution and weights and modifies only the assignment of delays to neurons, the degradation isolates how much the network relies on neurons carrying the longest or the shortest delays, respectively. Permuting the longest delays first degrades accuracy substantially in layers 1 and 2, for both the learned and fixed models (Fig.~\ref{fig:delays_spe}b), while permuting the shortest delays first yields a degradation of comparable magnitude (Fig.~\ref{fig:delays_spe}c). These observations show that while delay distributions are composed of a majority of short delays (more than half of the delay population is under $10$ time-steps in Fig.~\ref{fig:delays_spe}a) with a sparse minority of long delay neurons (less than $25\%$ of the delay population is over $20$ time-steps in Fig.~\ref{fig:delays_spe}a), both short and long time-scales are used by the networks. In layers 1 and 2, the learned-delays model also degrades more steeply than the fixed one (Fig.~\ref{fig:delays_spe}b,c), indicating that learning converges to a structure that relies more strongly on its specific delay values. Layer 3 behaves differently: its degradation stays below $\sim2\%$ even under full permutation, cutting its recurrence entirely costs little (dashed line, Fig.~\ref{fig:delays_spe}b), and the learned and fixed models nearly coincide. Finally, the network’s dependence on recurrent structure is concentrated in the shallower layers, where delay learning establishes a heterogeneous temporal organization that combines fast and slow dynamics. The deepest layer, in contrast, appears more independent of recurrent temporal structure, possibly because it operates on already integrated representations.
\\

\begin{figure}
  \centering
  \includegraphics[width=0.90\linewidth]{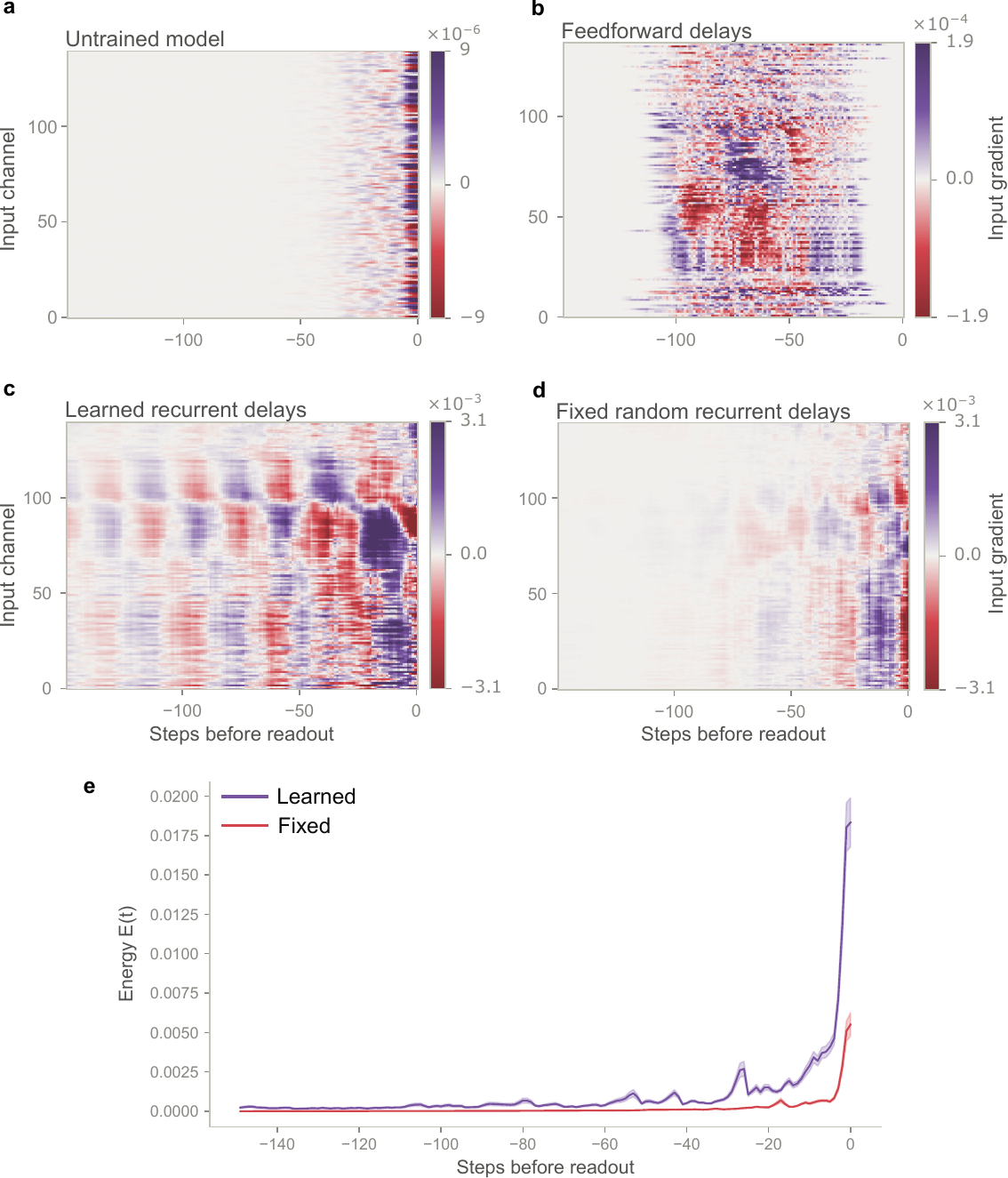}
  \caption{\textbf{Gradient maps reveal richer and longer-range temporal dependencies with learned recurrent delays.} \textbf{a-d}: Comparison of gradient maps for a model with untrained weights and delays (\textbf{a}), a model with only feedforward axonal delays trained as in \citep{Hammouamri2024} (\textbf{b}), a model with learned recurrent delays (\textbf{c}), and a control with fixed random recurrent delays (\textbf{d}). Panels show the gradients over a zero-input sequence of 3000 steps, and we display only the last 150 time-steps. In each case, the x-axis denotes time and the y-axis denotes input neuron index. The gradient is computed with respect to the activation of the output neuron corresponding to the target class, evaluated at the final time step of the zero-input sequence (the rightmost time-step in the plots). This example is taken for the class $4$ and seed $0$(gradient maps for all classes and seeds are available in the public repository). \textbf{e}: Energy profile of the gradient maps, averaged over all output classes and 5 seeds. Shaded areas represent SEM.}
  \label{fig:gradmaps}
\end{figure}

\noindent\textbf{Delay reorganization shapes the spatio-temporal selectivity.} Having established that delay optimization induces a structured reorganization of recurrent connectivity, we next asked whether this reorganization has a measurable consequence on the structure of the network's selectivity. To address this question, we analyzed gradient maps \citep{lindsey2019unifiedtheoryearlyvisual, ranconTemporalRecurrenceGeneral2025} with respect to the activation of a target output neuron at the final time step of a zero-input sequence. A gradient map quantifies the sensitivity of this output activation to infinitesimal perturbations of the input across input units and time, and thus provides a spatiotemporal visualization of how credit is assigned through the network (see Methods~\ref{sec:gradmap_method} for a more detailed explanation). Where Fig.~\ref{fig:delays_spe} characterizes the static organization of the trained recurrent parameters, Fig.~\ref{fig:gradmaps} probes their functional consequence for temporal dependencies. Fig.~\ref{fig:gradmaps}c shows that the learned-delays model develops higher amplitude gradients distributed over a broad portion of the sequence compared to the fixed random delays control which produces gradients that are weaker and concentrated near the latest time steps (Fig.~\ref{fig:gradmaps}c,d, shown on a common scale). This is not the same measurement as in Fig.~\ref{fig:gradflow}, which differs in the back-propagated scalar and the input(Methods~\ref{sec:gradmap_method}), so the two are not comparable quantitatively. Each isolates distinct effects: recurrent delays extend the effective temporal range of credit assignment (Fig.~\ref{fig:gradflow}b,c), and optimizing them restructures the back-propagated signal within that range.

The learned recurrent delays gradient maps (Fig.~\ref{fig:gradmaps}c) display clear neuron-specific and timing-specific temporal structure, while the fixed random delays control (Fig.~\ref{fig:gradmaps}d), despite an identical architecture, produces selectivity concentrated near the readout, indicating that a different spatio-temporal selectivity emerges with delay learning. For comparison, the untrained model (Fig.~\ref{fig:gradmaps}a) shows gradient patterns dominated by the final time steps with negligible amplitude, confirming that the temporal structure observed in trained models arises from learning rather than from the network architecture alone. The learned feedforward delays model (Fig.~\ref{fig:gradmaps}b) also displays a comparable long temporal sensitivity, yet with selectivity distributed over broader input neuron groups, and lacking the fine-grained temporal selectivity of the learned recurrent delays model (Fig.~\ref{fig:gradmaps}c).

To generalize this observation, we computed the energy profile (see Methods~\ref{sec:gradmap_method}) of the gradient maps, averaged over all output classes and the five seeds of trained models (Fig.~\ref{fig:gradmaps}e). We find that the learned-delay model maintains substantial gradient energy at intermediate and long lags, whereas the fixed-delay model is primarily sensitive to the final time steps before readout. These results show that optimizing recurrent delays extends the temporal range over which the network remains sensitive, enabling the capture of longer-range dependencies compared with fixed-delay networks (Fig.~\ref{fig:gradmaps}d). 

Overall, these results highlight two distinct contributions of recurrent delays. First, their presence extends the temporal range over which credit is assigned, independently of whether delays are optimized. Second, recurrent delay learning reshapes the network’s spatio-temporal selectivity into a structured organization that is functionally exploited by the network, enabling richer and longer-range temporal processing than weight optimization alone at comparable accuracy.

\subsection{Recurrent delay learning maintains performance under constraints} \label{sec:constraints}

The benchmarks in Section~\ref{sec:tasks} establish the overall benefits of incorporating recurrent delays across a range of temporal tasks. Notably, heterogeneous recurrent delays remain highly effective even when kept fixed: a well-initialized fixed-delay network already matches the performance of substantially more complex neuron models (see Table~\ref{tab:SSC-PSMNIST} and Table~\ref{tab:AL-HAR}), which is a result of independent interest for neuromorphic systems, where a random but well-chosen delay distribution is training-free yet buys substantial temporal expressivity. However, this observation indicates that accuracy metrics on standard benchmarks may capture only a part of what delay learning provides. In this section, we move beyond accuracy as the sole target and study the benefits of our delay learning algorithm under two families of constraints directly relevant to neuromorphic deployment: efficiency and robustness, where robustness refers to maintaining stable performance under degraded or perturbed inputs. We run these analyses on the HAR dataset, which is both strongly temporal \citep{chenNeuromorphicSequentialArena2025} and small enough to facilitate large sweeps over initialization, regularization strength and input perturbation across multiple seeds.

The delay–weight co-organization directly shapes the two dominant costs of running an RSNN on neuromorphic hardware: the memory needed to buffer future spikes, set by the maximum delay, and the energy spent emitting them, set by the firing rate. Biological circuits share the same constraints in axonal length and myelination for delays \citep{Monje2018}, and in metabolic cost for spikes. In either substrate, reaching a given accuracy with shorter delays and fewer spikes is desirable. 
\\

\begin{figure}
\centering
\includegraphics[width=\textwidth]{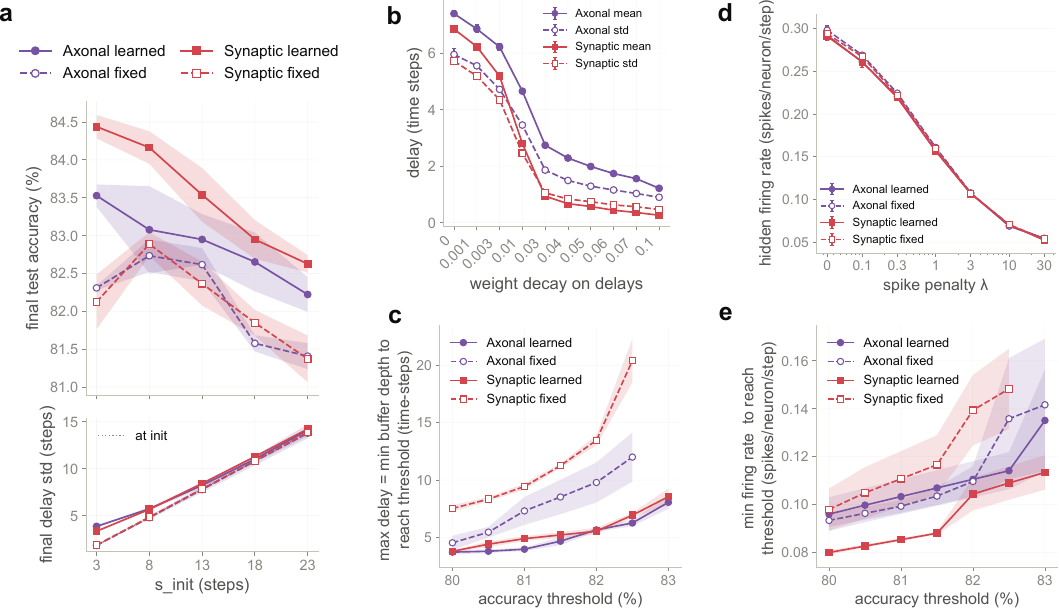}
\caption{\textbf{Learned delays enable more energy and memory efficient networks.} \textbf{a}, Top: final test accuracy versus the scale $s_{\text{init}}$ of the half-normal recurrent-delay initialization, for axonal/synaptic and learned/fixed variants. Bottom: final delay std versus its initialization. The dotted line is the delay spread realized at initialization, which the frozen fixed families sit on by construction. \textbf{b}, Mean and std of the learned delays as a function of the weight-decay strength applied to the delay parameters: increasing the penalty compresses the delays toward zero. \textbf{c}, Minimum buffer depth (equivalently, maximum delay) required to reach a given accuracy threshold. At every threshold, learned delays need a shorter maximum delay than fixed ones. Learned delays are compressed by weight decay on the delay parameters, fixed delays by initializing at a smaller $s_{\text{init}}$. A curve stops at the first threshold one of its 3 seeds never reaches. \textbf{d}, Hidden-layers firing rate as a function of the spike-count penalty $\lambda$ added to the loss. \textbf{e}, Minimum firing rate required to reach a given accuracy threshold. Learned synaptic delays reach each threshold at a lower firing rate than their fixed control; for axonal delays, the two are indistinguishable. Curves show mean $\pm$ SEM over 3 seeds: in \textbf{c},\textbf{e} the smallest cost reaching the threshold is read separately for each seed, interpolating between the two configurations on either side of it, and then averaged.}
\label{fig:constraints}
\end{figure}

\noindent\textbf{Delay learning can reduce the network's memory footprint}. We first investigate whether training alone shortens the delays. Varying the scale $s_{\text{init}}$ of the half-normal recurrent-delay initialization (Fig.~\ref{fig:constraints}), fixed-delay networks are penalized at the shortest initializations and need a sufficient delay spread to reach their best accuracy. However, a wider spread means a larger maximum delay, and therefore a deeper buffer. While learned delays outperform their fixed counterparts across the whole range, the final delay spread remains mostly governed by its initialization (Fig.~\ref{fig:constraints}a, bottom): optimization tends to widen the spread, especially for short initializations, but does not reorganize it enough to decouple the final buffer depth from the initial one. To apply an explicit pressure, we add a weight-decay term on the delay parameters to mechanically pull both their mean and their standard deviation toward zero as its strength increases (Fig.~\ref{fig:constraints}b). For fixed delays, which are not trained, we instead compress the distribution manually, by sweeping the half-gaussian initialization toward smaller standard deviations (Methods~\ref{sec:methods_constraints}). We then report the minimum buffer depth (or equivalently the maximum delay) required to reach each accuracy threshold (Fig.~\ref{fig:constraints}c). Learned delays meet every plotted threshold with a shorter maximum delay than the fixed baseline, so at matched accuracy, delay learning uses shorter delays, and thus less memory. 
\\

\noindent\textbf{Delay learning can reduce the network's energy consumption}. On event-driven neuromorphic chips, energy consumption depends on the spike counts. Therefore, to reduce energy consumption, we can add a spike-count penalty $\lambda$ to the loss and lower the firing rate of the network (Fig.~\ref{fig:constraints}d), thereby enabling a trade-off between neural activity and accuracy. Comparing the minimum firing rate required to achieve each accuracy threshold (Fig.~\ref{fig:constraints}e), learned synaptic delays consistently require fewer spikes than fixed delays. For axonal delays, the two curves are indistinguishable, and learning buys no spike saving. This parallels the accuracy results, where synaptic delays are the stronger configuration on HAR (Table~\ref{tab:ablation}), suggesting that the finer temporal structure of per-synapse delays is what translates into an energy saving, while the coarser per-neuron delays do not. Both compression sweeps and the spike-penalty sweep use the delay initialization most favorable to the fixed baseline ($s_{\text{init}} = 8$, Fig.~\ref{fig:constraints}b-e), so that every comparison is conservative for the learned models. Taken together, these results show that learned recurrent delays reduce the memory and energy cost of achieving a given level of accuracy. Compared with fixed delays, they require shallower delay buffers and, for synaptic delays, lower firing rates to attain the same performance. The advantage of delay learning is thus amplified once efficiency, rather than accuracy alone, is taken as the objective.
\\

\begin{figure}
\centering
\includegraphics[width=\textwidth]{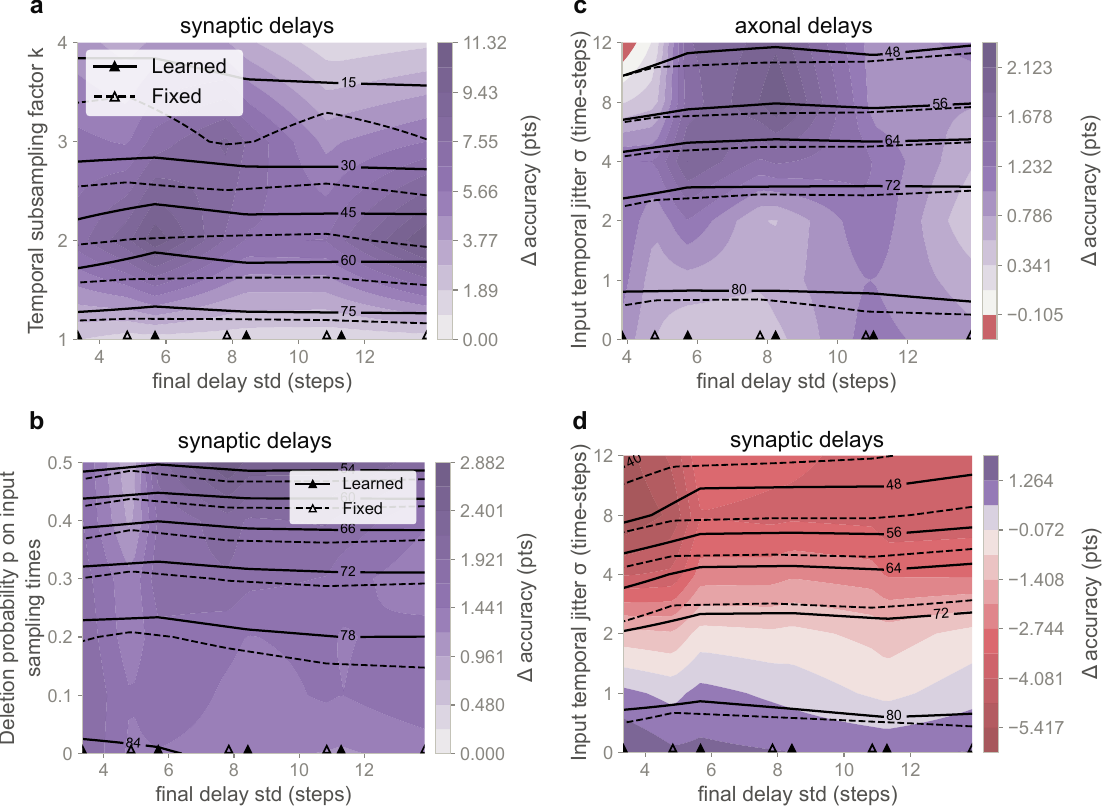}
\caption{\textbf{Robustness of learned versus fixed recurrent delays to temporal perturbations.} Color shows the accuracy difference between learned and fixed delays ($\Delta$ accuracy $=$ learned $-$ fixed) as a function of perturbation strength (y-axis) and final recurrent-delay spread (x-axis). Solid and dashed lines are the iso-accuracy contours of the learned and fixed models, respectively. Models come from the initialization sweep ($s_{\text{init}} \in \{3,8,13,18,23\}$), and filled and open triangles mark the final delay std values reached by learned and fixed models, respectively. \textbf{a}, Temporal subsampling by factor $k$ (synaptic). \textbf{b}, Random input-sample deletion with probability $p$ (synaptic). \textbf{c}, Gaussian jitter $\sigma$ on input sampling times (axonal). \textbf{d}, The same jitter (synaptic). Mean over $3$ model seeds for \textbf{a}, and over 3 model seeds × 3 perturbation realizations for \textbf{b–d}.}
\label{fig:robustness}
\end{figure}

\noindent\textbf{Delay learning shapes the robustness to data perturbation.} Alongside efficiency, a second requirement for reliable hardware deployment is robustness under noise or data corruption. We evaluated this aspect in learned and fixed delays networks by probing their accuracy on the HAR test set while applying three temporal perturbations to the inputs: temporal subsampling by a factor $k$, in which only every $k$-th time step is kept and the intermediate ones are set to zero, random deletion of individual input time-steps with probability $p$, and a random gaussian temporal jitter $\sigma$ on the input samples. To compare learned and fixed models at approximately matched delay distributions rather than at matched initialization, we interpolate each accuracy map across the final delay spread reached by every network (Fig.~\ref{fig:robustness}), so that a given point on the x-axis corresponds to a similar delay distribution in both models. For subsampling and deletion, we show the synaptic case only, the axonal one being qualitatively identical (Supplementary~\ref{sup:perturbations})

Under subsampling (Fig.~\ref{fig:robustness}a) and deletion (Fig.~\ref{fig:robustness}b), learned delays systematically keep a positive advantage over fixed ones across the whole range of delay distributions and perturbation magnitudes. However, under temporal jitter on the inputs, the robustness offered by delay learning depends on the delay type: for axonal delays learning keeps its advantage over fixed ones (Fig.~\ref{fig:robustness}c), but this advantage reverses for synaptic delays, where the learned model is much more sensitive to the input timing perturbation than its fixed counterpart (Fig.~\ref{fig:robustness}d, negative $\Delta$ accuracy). We hypothesize that these observations reflect both the nature of the perturbations and the difference in encoding between axonal and synaptic delays. Subsampling and deletion mostly change which events reach the network, whereas jitter changes when each event arrives. If this distinction drives the results, then the greater sensitivity of synaptic delays to jitter would reveal that their encoding differs from the axonal one in being more finely tuned to the exact input latencies. This interpretation echoes the energy saving results: the fine per-synapse timing structure that converts into an energy saving (Fig.~\ref{fig:constraints}e) may also be what makes the network most sensitive to input timing noise.

These results show that learned recurrent delays can improve both the efficiency and the robustness of the network beyond what accuracy alone reveals. Delay learning may thus be valuable not only to improve the performance of models, but also to shape how that performance is achieved, with the possibility of leveraging different delay types and regimes in order to satisfy deployment constraints and temporal demands of a given task. 

\section{Discussion}\label{sec:discussion}

Conduction delays are a fundamental feature of biological neural circuits. However, in most existing RSNNs, recurrent delays are either omitted entirely or treated as fixed architectural parameters, as optimizing them remains technically challenging. This work addresses this gap by introducing DelRec, a new method for optimizing recurrent delays using surrogate gradient learning and backpropagation. Compatible with any spiking neuron model, DelRec only requires a small modification, adding learnable delay parameters at the neuron or synapse level, making delay plasticity readily applicable to existing recurrent SNN architectures. The performance of this approach on temporal benchmarks demonstrates that temporal expressivity in RSNNs can be obtained without elaborate neuron models, but can instead emerge from the structure of recurrent interactions. Moreover, our results highlight two distinct contributions of recurrent delays. First, the presence of recurrent delays extends the temporal range over which credit is assigned and enriches the model's expressivity, even when the delays are not optimized, leading to substantial performance gains across benchmarks. We then show that optimizing recurrent delays reorganizes the network's spatio-temporal selectivity into a structured representation that the network actively exploits, and that this reorganization enables more memory- and energy-efficient models.

The accessibility to such a method creates new opportunities for computational modeling of biological networks, where recurrent (or horizontal) connections are thought to contribute to a range of important phenomena, from traveling waves \citep{davisSpontaneousTravelingWaves2021} to polychronization \citep{izhikevichPolychronizationComputationSpikes2006}. By enabling the simulation and optimization of these parameters, DelRec provides a new avenue for studying how delay-dependent dynamics and computations emerge in neural populations. As highlighted in this work, delay optimization does not simply increase delay values but induces a structured self-organization of temporal dynamics. Depth-dependent distributions emerge across layers (Fig.~\ref{fig:delays_spe}), while weight and delay co-adaptation promotes neuron specialization within layers beyond what can be achieved through weight learning alone. Such layer-specific temporal specialization is reminiscent of the hierarchical timescale organization observed along cortical processing streams \citep{Murray2014Hierarchy} (see Section~\ref{self orga}). Whether similar temporal structures emerge across different tasks, architectures, or neuron models remains an open question that can now be systematically investigated.

Beyond these modeling insights, recurrent delay optimization has direct consequences for neuromorphic engineering. This work demonstrates that delays provide an efficient mechanism for shaping recurrent dynamics in a spiking neural network without increasing the complexity of the neuron model itself. Indeed, DelRec achieves state-of-the-art accuracy on multiple established benchmarks for temporal processing, using simple LIF neurons. We show that recurrent delay learning enables competitive temporal processing while reducing network memory footprint and energy consumption. This finding is of practical importance for neuromorphic hardware, as programmable delays are already available on many existing platforms (e.g., Intel Loihi, IBM TrueNorth, SpiNNaker, SENeCA) while incurring only modest energy overheads, estimated at around $10\%$ on Loihi \citep{meszaros2025completepipelinedeployingsnns}. DelRec provides a powerful method to determine optimal delay configurations during training while satisfying memory and sparsity constraints, which could directly influence the design of neuromorphic processors.

Several limitations should nevertheless be noted. DelRec introduces additional computational overhead during training, primarily due to the spike scheduling mechanism and temporal spread annealing. However, this overhead can be largely mitigated by implementing recurrent delays as custom GPU kernels instead of relying on a standard automatic differentiation graph.  We provide such an implementation in Triton (Supplementary~\ref{computational_overhead}) in the public repository, reducing the training time and memory footprint of recurrent layers with learned delays to levels close to those of conventional RSNNs in practice. In addition, recent studies suggest that the performance gains from delay optimization in spiking neural networks are often modest on some datasets relative to fixed-delay models \citep{Hammouamri2024, karilanova2026delaysspikingneuralnetworks, vassallo2026factordelaylearningrules, delay_gradient_structure}. Our gains are also relatively modest, depending on the temporality of the dataset and the quality of the delay initialization. This may indicate that the benchmark is not sufficiently temporally demanding to fully exploit the additional temporal capacity provided by learned delays, making this extra parameterization unnecessary. However, these results may also reflect interactions between synaptic weights and delays during optimization that are not yet well understood. Characterizing these interactions and determining when they become beneficial or detrimental remains an important avenue for future research. Nevertheless, our results show that delay learning is not only a mechanism for improving task performance, but also a means of shaping the representations through which this performance is achieved. This new perspective may offer a more informative way of assessing the benefits of delay optimization than performance alone.

To conclude, our findings suggest that the temporal structure of recurrent interactions represents a powerful and largely untapped computational dimension in spiking neural networks. By enabling its optimization, DelRec opens new avenues for designing more efficient neuromorphic systems and for investigating the role of delay plasticity in biological computation.

\section{Methods}\label{sec11}

\subsection{Detailed delay learning procedure} \label{sec:learning_procedure}

We detail here the scheduling mechanism introduced in Section~\ref{sec:learning-delays} and its memory footprint. Consider one layer of $N$ different neurons, with input sequences of temporal dimension $T$. Whenever neuron $j$ emits a spike at time $t$, we schedule an input $w_{ij}^{\text{rec}} S_j[t]$ at time $t + 1 + d_j$ for all neurons $i$ of the same layer, spread over the neighboring time steps by the kernel $h_{\sigma, d_j}$ of Eq.~\ref{eq:10}. This is written for axonal delays, where $d_j$ is carried by the presynaptic neuron $j$, but for synaptic delays $d_j$ is replaced by $d_{ij}$ and the kernel is evaluated per connection. In this framework, the learnable parameters are the recurrent delays $d$ and the synaptic weights $w$.

The function $h_{\sigma, d}(\tau)$ has a finite support $\text{supp}(h_{\sigma, d}(\tau))$, depending only on $\sigma$ and $d$. Indeed:
\begin{align}
    \forall \tau, \; h_{\sigma, d}(\tau) \neq 0 \Leftrightarrow
    \tau \in \text{supp}(h_{\sigma, d})
    &= \big[(1+d) - (1+\sigma) \; ; \; (1+d) + (1+\sigma)\big] \notag\\
    &= \big[d - \sigma \; ; \; 2 + d + \sigma\big]. \label{eq:supp_h}
\end{align}

Consequently, it is sufficient to schedule recurrent inputs only within the support of $\text{supp}(h_{\sigma, d})$, and as $\sigma$ decreases during training, the range of time steps when we can schedule recurrent inputs becomes narrower. At the scale of a layer, we need to schedule inputs for multiple neurons at the same time, which means it suffices to compute and schedule inputs for a limited range of time steps $\textbf{E}$, such that:
$$\tau \in \textbf{E}(\sigma, D) = \bigcup_{d\; \in \;  D}\text{supp}(h_{\sigma, d}),$$
with $D = \{d_j \; ; \; j = 1, \; ... \; , N\}$ the delays of the $N$ neurons of the layer. In practice, as $\sigma$ is decreasing to $0$ in Eq.~\ref{eq:supp_h}, we ignore the lower bound of $\textbf{E}$, and we approximate this set with:
\begin{equation}
\tilde{\mathbf{E}}(\sigma, D)
   = \bigl[\, 0 \,;\; 
      \, \bigl\lceil\, 1 + \max_{1 \leq j \leq N} d_j + (1+\sigma) \,\bigr\rceil
     \,\bigr] \supset \textbf{E}(\sigma, D). \label{eq : define-E-tilde}
\end{equation}

In other words, we only need to compute and store $h_{\sigma, d}(\tau)$ for $\tau$ in $\tilde{\mathbf{E}}(\sigma, D)$, so the scheduling matrix $X^{rec}$ has in fact a dimension of $N \times \text{dim}(\tilde{\mathbf{E}}(\sigma, D))$, then at time $t$, the future recurrent input at time $t + \tau$ is updated with Eq.~\ref{eq:rec_update}. We use $X^\mathrm{rec}$ as a ring buffer with a pointer mechanism in order to efficiently schedule the future recurrent inputs (see Algorithm~\ref{alg:learn-delays}). Finally, the delays are constrained to stay positive.

\subsection{Detailed forward pass algorithm}

We detail here the algorithm we use to schedule delays in recurrent connections. Algorithm~\ref{alg:learn-delays} describes a forward pass for $N$ neurons. Our spike scheduling employs a circular buffer of size $N \times \text{dim}(\tilde{\mathbf{E}}(\sigma, D))$ with $\tilde{\mathbf{E}}(\sigma, D)$ defined in Eq.~\ref{eq : define-E-tilde}. 

\begin{algorithm}[t] 
\caption{Scheduling delays in Recurrent Connections}
\label{alg:learn-delays}
\begin{algorithmic}[1]
\Input Sequence time point at time $t$: $X[t]$, for $N$ neurons
\Param Number of neurons $N$, recurrent weights $W^{rec}\in\mathbb{R}^{N\times N}$, recurrent delays $d\in\mathbb{R}^{N}$, schedule $\sigma_{epoch}$, firing threshold $V_{th}$, neuron reset type, reset potential $V_{reset}$, neuronal charge function $f$
\Output Spikes at time $t$: $S[t]$
\State Compute $\tilde{E}(\sigma_{epoch}, D)$
\State $L \gets \dim\!\big(\tilde{E}(\sigma_{epoch}, D)\big)$
\State $\mathrm{spread}[i] \gets \big[\, h_{\sigma_{epoch},\, d_i}(x) \ \text{for}\ x \in \tilde{E}(\sigma_{epoch}, D) \,\big], \quad i = 1,\dots,N$ 
\State $B \gets 0^{N\times L}$ \Comment{zero-out buffer matrix}
\State $\mathrm{pointer} \gets 0$
\For{$t \gets 0$ \textbf{to} $T$}
    \For{$i \gets 1$ \textbf{to} $N$}
        \State $X^{rec}_i[t] \gets B_i[\mathrm{pointer}]$ \Comment{take current time step in the buffer}
        \State $I_i[t] \gets X_i[t] + X^{rec}_i[t]$
        \State $H_i[t] \gets f\big(V_i[t-1], I_i[t]\big)$
        \State $S_i[t] \gets \Theta\big(H_i[t] - V_{th}\big)$
        \If{hard reset}
            \State $V_i[t] \gets H_i[t]\,(1 - S_i[t]) + V_{reset}\, S_i[t]$
        \Else
            \State $V_i[t] \gets H_i[t] - V_{th}\, S_i[t]$
        \EndIf
        \State $B_i[\mathrm{pointer}] \gets 0$ \Comment{clear the slot just read}
        \For{$j \gets 1$ \textbf{to} $N$}
            \For{\textbf{each} $\tau \in \tilde{E}(\sigma_{epoch}, D)$ \textbf{with} $\tau \geq 1$} 
                \If{$\mathrm{pointer} + \tau \leq L - 1$}
                    \State $B_j[\mathrm{pointer} + \tau] \gets B_j[\mathrm{pointer} + \tau] + w^{rec}_{ji}\cdot \mathrm{spread}[i][\tau]\cdot S_i[t]$
                \Else
                    \State $p' \gets (\mathrm{pointer} + \tau) \bmod L$
                    \State $B_j[p'] \gets B_j[p'] + w^{rec}_{ji}\cdot \mathrm{spread}[i][\tau]\cdot S_i[t]$
                \EndIf
            \EndFor
        \EndFor
    \EndFor
    \State $\mathrm{pointer} \gets (\mathrm{pointer} + 1) \bmod L$ \Comment{advance once per time step, after scheduling}
\EndFor
\end{algorithmic}
\end{algorithm}

\subsection{Mackey-Glass time-series prediction} \label{mg_methods}

\noindent\textbf{Task.} Time series are generated by numerically integrating Eq.~\ref{eq:mackey-glass} with
parameters $\beta = 0.2$, $\gamma = 0.1$, $n = 10$, and initial condition $x_0 = 1.2$,
using a forward Euler scheme with step size $\mathrm{d}t = 0.1$. The resulting continuous
trajectory is subsampled at unit intervals (i.e.,\ every $1/\mathrm{d}t = 10$ integration
steps). The first 5000 time steps after subsampling are discarded to eliminate transient
dynamics, yielding a stationary series of 6000 points. This series is split chronologically
into training ($60\%$), validation ($20\%$), and test ($20\%$) sets, so that no time step is
shared across splits. All splits are standardized to zero mean and unit variance using the
training set statistics. We sweep over the system delay $\tau \in \{17, 30, 40, 50\}$ and the
prediction horizon $H \in \{20, 30, 40, 50\}$, generating an independent time series for each
value of $\tau$. Every $(\tau, H)$ condition is trained over three random seeds. The network is trained on windows of $W=150$ time-steps ($x(t), x(t+1), \dots, x(t+W-1)$) to
predict the value $x(t+W-1+H)$ by minimizing the mean squared error (MSE). In
Fig.~\ref{fig:mackey-glass}, we report the normalized MSE (NMSE), defined as the MSE divided
by the variance of the target values on the test set. The NMSE is thus the fraction of the
signal's variability that the model fails to explain, which allows the error to be compared across all $\tau$ values.
\\

\noindent\textbf{Compared models.} All four variants share the same architecture: a single feedforward projection from the input
(dimension 1) to a single hidden layer of $128$ recurrently connected neurons, followed by a
linear readout to a single output neuron. LIF neurons use a membrane time-constant of
$\tau_{\text{membrane}}=2$ (in time-steps), $V_{\text{threshold}}=1$, $V_{\text{reset}}=0$ and
the arctan surrogate gradient with $\alpha = 5$. The readout integrates the hidden spike-trains through a leaky-integrator neuron with time constant $\tau_{\text{readout}} = 20$ (an exponential moving average with decay $1-1/\tau_{\text{readout}}$), and its final membrane voltage is taken as the prediction. The readout time-constant is set larger than that of the hidden neurons so that the readout integrates recurrent spike trains over larger temporal windows, producing a smooth prediction, whereas the hidden neurons can encode fine temporal changes. The four variants differ only in their recurrent delays which, when present, are axonal (one learnable delay per presynaptic neuron). No delays: a plain RSNN whose recurrent delays are frozen at zero. Fixed delays: recurrent delays frozen at a random initialization drawn uniformly on $[0, 20]$ time-steps, the control that isolates the contribution of delay learning. Learned delays without $\sigma$-annealing: the same delays trained jointly with the weights, but with the triangle interpolation width held at $\sigma = 0$ throughout. DelRec: learned delays with the $\sigma$-annealing schedule of Section~\ref{sec:learning-delays}, with $\sigma$ initialized at $10$ and decayed by a factor $0.95$ at each epoch. Weights and delays are optimized by two separate Adam optimizers (learning rate $5\times10^{-4}$ with weight decay $10^{-4}$ for the weights, and $10^{-1}$ for the delays), each following a cosine schedule, and all models are trained for $100$ epochs with a batch size of $512$. Delays are rounded to the nearest integers at each epoch. For each run, we retain the checkpoint with the lowest validation NMSE and evaluate it on all three splits with $\sigma$ set to $0$. All remaining hyperparameters are given in the Mackey-Glass script available at \url{https://github.com/alexmaxad/DelRec}.

\subsection{Temporal classification benchmarks} \label{sec:temporal benchmarks}

\noindent\textbf{Tasks.} We evaluate DelRec on four classification benchmarks. The implementations for SSC and PS-MNIST build on the codebase of \citep{xuASRCSNNAdaptiveSkip2025}, while AL and HAR follow the benchmark of \citep{chenNeuromorphicSequentialArena2025}, which evaluates every model within a single shared pipeline at a matched parameter budget of $\approx 100$k, so that differences in accuracy reflect the temporal-processing mechanism rather than model size or training setup. Across all four benchmarks, the networks use plain LIF neurons and fully connected recurrent hidden layers, with neither normalization layers nor data augmentation. To isolate the contribution of delay learning, every experiment compares learned delays against randomly initialized delays frozen at their initialization (fixed). We further compare axonal (one delay per neuron) with synaptic (one delay per connection) delays. Reported accuracies are means ($\pm$ std) over five random seeds for SSC, PS-MNIST and AL. The scores for the HAR dataset are achieved by the best configuration for each model type from the sweep done on the standard deviation of the delay distribution in Section~\ref{sec:constraints}.
\\

\noindent\textbf{Readout and loss.} The readout is a single linear layer mapping the last hidden layer to the class logits through a layer of leaky-integrate neurons, and its membrane-voltage sequence $V^{\text{out}}[t]$ is reduced over time and passed to a cross-entropy loss. The reduction is a sum over time steps for SSC and PS-MNIST, and a temporal average for AL and HAR, matching the readout of \citep{chenNeuromorphicSequentialArena2025}. Denoting by $o_n = \textstyle\sum_t\, V^{\text{out}}[t] \in \mathbb{R}^{n_{\text{classes}}}$ the reduced logits of sample $n$ in a batch of size $N$, and $y_n$ its label, the loss is:
\begin{equation}
    \mathcal{L} = -\frac{1}{N} \sum_{n=1}^{N} \log\!\Big( \operatorname{softmax}(o_n)_{y_n} \Big).
\end{equation}

\noindent\textbf{Sigma scheduling.} The delays are learned jointly with the weights following the procedure of Methods~\ref{sec:learning_procedure}: a spike is scheduled onto its target neurons through a triangular spread $h_{\sigma, d}$ centered on the delay $d$, whose width $\sigma_{epoch}$ is annealed from an initial value down to zero over training so that, by the end, each connection carries a single sharp delay. We update $\sigma_{epoch}$ at each epoch with an exponential decay ($decay = 0.95$),
\begin{equation}
    \sigma_{epoch} \leftarrow \sigma_{init} \times decay^{\,100 \times \frac{epoch}{N_\text{epochs}}}, \label{eq:update_sigma}
\end{equation}
$N_\text{epochs}$ being the total number of epochs, $\sigma_{init}$ is set per benchmark (Supplementary~\ref{hyperparameters}), and we found that using a $\sigma_{init}>0$ did not help on AL and HAR, so we fixed it at $0$. On SSC and PS-MNIST we additionally allow the spread to narrow at a different rate for each neuron: we introduce a per-neuron parameter $p_i$ (initialized at $0$ and added to the learnable parameters) and use for neuron $i$ the modified spread:
\begin{equation}
    h_{\sigma, d_i, p_i}(\tau) = \max\!\left(0,\; \frac{ 1 + 2\cdot \text{sig}(p_i) \cdot \sigma - |\tau - (1 + d_i)| }{ (1 + 2\cdot \text{sig}(p_i) \cdot \sigma)^2 } \right),
    \label{eq:spread_p}
\end{equation}
with $\text{sig}(\cdot)$ the sigmoid function. 
\\

\noindent\textbf{Model selection on AL and HAR.} Neither dataset provides a
validation split, so we follow \citep{chenNeuromorphicSequentialArena2025} and report the accuracy of the best checkpoint over the training epochs. On HAR, each delay family is reported at the initialization scale $s_{\text{init}} \in \{3, 8, 13, 18, 23\}$ maximizing its accuracy in the sweep of Section~\ref{sec:constraints}.
\\

\noindent\textbf{Delays at inference.} To preserve compatibility with neuromorphic hardware, learned delays are rounded to the nearest integer at inference, leaving a single integer delay per connection. This discretization has little effect on the SSC, PS-MNIST, and HAR datasets, but substantially affects performance on the AL dataset, as detailed in Supplementary~\ref{delay rounding effect}. Accordingly, the results on the AL dataset are reported with $\sigma=0$, where delay values are represented as a weighted combination of the two nearest integers rather than being rounded.
\\

\noindent\textbf{More complex neuron models.} We deliberately leave out of Table~\ref{tab:SSC-PSMNIST} models that rely on substantially more complex neuron models, such as multi-compartment neurons \citep{zheng2024temporal, chen2024pmsn}, attention or GRU based neurons \citep{dampfhoffer2022investigating, wang2024efficientspeechcommandrecognition}, whose additional mechanisms make direct comparison less meaningful\footnote{\citep{dampfhoffer2022investigating} report 77$\pm$0.4\% on SSC using a spiking GRU, \citep{zheng2024temporal} report 82.46\% on SSC (on only one seed) using multi-compartment neurons, \citep{wang2024efficientspeechcommandrecognition} report 83.69\% on SSC (only one seed) using attention and distillation, \citep{chen2024pmsn} report 97.78\% on the PS-MNIST (one seed) with a multi-compartment neuron and using the test set as the validation set.}.

\subsection{Gradient propagation during training} \label{sec:methods_gradflow}

To probe how far back in time error signals survive, we logged throughout training the norm of the gradient of a probe loss (defined below) $\mathcal{L}_{\text{probe}}$ with respect to the network input at each past time step,
\begin{equation}
    g(t) \;=\; \left\lVert \frac{\partial \mathcal{L}_{\text{probe}}}{\partial x[t]} \right\rVert
    \;=\; \left( \sum_{n=1}^{N_{\text{in}}} \left( \frac{\partial \mathcal{L}_{\text{probe}}}{\partial x[t,n]} \right)^{\!2} \right)^{\!1/2},
    \label{eq:gradflow}
\end{equation}
where $x[t,n]$ is the input at time step $t$ in channel $n$, and the $L_2$ norm is taken over the $N_{\text{in}} = 140$ SSC input channels and then averaged over the samples of the probe batch. In Fig.~\ref{fig:gradflow} we report $g$ as a function of the temporal depth $k = (T-1) - t$, i.e.,\ the number of time steps separating that input from the readout ($T = 250$ steps of $5.6$\,ms), and of the training epoch.
\\

\noindent\textbf{Probe loss.} The training loss $\mathcal{L}$ of Methods~\ref{sec:temporal benchmarks} cannot serve as the probe. It reduces the readout by summing $V^{\text{out}}[t]$ over all time steps, so every input time step has a direct feedforward path to the loss. Computing $g$ for this loss would therefore not isolate the propagation of gradients through the recurrence. We therefore back-propagate the cross-entropy of the final time-step readout alone,
\begin{equation}
    \mathcal{L}_{\text{probe}} \;=\; -\log\!\Big( \operatorname{softmax}\big(V^{\text{out}}[T-1]\big)_{y} \Big),
\end{equation}
so that no input time step retains a direct path to the loss: what reaches an input at temporal depth $k$ has necessarily propagated back through the network's own recurrent dynamics. This matches the loss point used for the gradient maps of Methods~\ref{sec:gradmap_method}, which are likewise evaluated at the last time step of the sequence. Gradients are propagated through the same Triangle surrogate used for training.
\\

\noindent\textbf{Probed batch.} The probe is a separate forward and backward pass performed once per epoch, immediately after that epoch's weight and delay updates. It does not contribute to training. It uses a single fixed, class-balanced batch of $245$ held-out samples ($7$ per class $\times$ $35$ classes), drawn once from the SSC test set with a fixed random seed and kept identical across all epochs and all models, so that the rows of Fig.~\ref{fig:gradflow} and the three panels are directly comparable. During the probe, dropout is disabled and the delay spread $\sigma_{epoch}$ is left at its current annealed value, so the network is probed exactly in the state in which it is being trained.
\\

\noindent\textbf{Compared models.} Three networks are compared, sharing the SSC architecture, hyperparameters and random seed of Methods~\ref{sec:temporal benchmarks}, and therefore the same weight initialization and the same initial draw of the recurrent delays: a standard RSNN (all recurrent delays frozen at $d=0$), a network with axonal recurrent delays frozen at their random initialization, and one with learned axonal recurrent delays. The standard RSNN required a lower weight learning rate ($2\times10^{-4}$ instead of $10^{-3}$) to train stably. Only the learning rate was changed, so the presence of delays remains the single architectural difference between the three models.

\subsection{Delay distributions and permutation ablation}
\label{sec:methods_permutation}

Both analyses of Fig.~\ref{fig:delays_spe} are performed on the axonal SSC models reported in Table~\ref{tab:ablation}, using the best validation checkpoint from each run, evaluated according to the protocol described in Methods~\ref{sec:temporal benchmarks} ($\sigma = 0$ and integer delays), with $5$ seeds per family.
\\

\noindent\textbf{Delay distributions (Fig.~\ref{fig:delays_spe}a).} For each hidden layer, we computed the empirical cumulative distribution of the delay distribution after training. The curves are averaged across $5$ seeds and the shaded bands provide the standard deviation across seeds. The strip below the axis reports the mean $\pm$ standard deviation of the delay distribution for each layer, averaged across seeds. Delays are reported as the raw parameter $d$, while the physical lag is $1 + d$ time steps (Methods.~\ref{sec:learning_procedure}). The initialization reference ("init") is the fixed-delays model, whose delays are frozen at the same half-normal draw ($s_{\text{init}}=12$, Supplementary~\ref{hyperparameters}) that initializes the learned runs (its three layers are statistically indistinguishable and are therefore pooled into a single curve). For display, the delay axis is cropped at the $99.5$th percentile of the pooled delays: the remaining $0.5\%$ is the sparse long-delay tail, which extends to $59$ time steps.
\\

\noindent\textbf{Permutation ablation (Fig.~\ref{fig:delays_spe}b,c).} Starting from a trained network, we aimed to quantify how much each layer's function depends on its recurrent delay distribution. For one recurrent layer at a time and a fraction $q$, we ranked the layer's $N$ delays and selected either the $qN$ longest (Fig.~\ref{fig:delays_spe}b) or the $qN$ shortest (Fig.~\ref{fig:delays_spe}c), then randomly permuted their values among the selected neurons. All recurrent and feedforward weights, all biases, and the delays of every other layer are left untouched, and the perturbed delays remain integer-valued. The network is then re-evaluated on the clean test set. We swept $q \in \{0, 0.1, 0.25, 0.5, 0.75, 1\}$, $q=0$ being the unperturbed network and $q=1$ a full shuffle of the layer. Because a permutation leaves the layer's marginal delay distribution, its weights and its total recurrent drive unchanged, it allows to precisely target the importance of the delay-weight structure in the permuted portion of the distribution. For each model, we report the relative degradation in test accuracy relative to the unpermuted trained network,
\begin{equation}
    \Delta_{\text{rel}}(q) \;=\; 100 \times \frac{a(0) - a(q)}{a(0)},
\end{equation}
where $a(q)$ is the test accuracy after the lesion and $a(0)$ that of the same run before it. Each point is the mean over $5$ model seeds $\times$ $3$ independent permutation draws, and the band is the SEM over those $15$ evaluations. The dashed horizontal line in each panel indicates the cost of removing recurrence from that layer by setting its entire recurrent weight matrix to zero. Values are averaged across random seeds and across both the learned- and fixed-delay model families. 

\subsection{Gradient maps} \label{sec:gradmap_method}

\noindent\textbf{Selectivity maps.} Methods~\ref{sec:methods_gradflow} quantifies, during training and on real input sequences, how far back in time the gradient propagates. To characterize the resulting spatio-temporal selectivity across model variants, we use gradient maps \citep{lindsey2019unifiedtheoryearlyvisual, ranconTemporalRecurrenceGeneral2025}, which computes the sensitivity of a model's chosen output to infinitesimal perturbations in the inputs across neurons and time (also called the spatio-temporal receptive field). Consider a trained network with $N_{\text{in}}$ input neurons, and an input time series of $T$ time-steps, and let $V^{\text{out}}_c[T-1]$ denote the readout membrane potential of class $c$ at the final time step. The gradient map of class $c$ is the matrix:
\begin{equation}
    G_c[t,n] \;=\; \left. \frac{\partial\, V^{\text{out}}_c[T-1]}{\partial\, x[t,n]} \right|_{x = 0},
\end{equation}
where $x[t,n]$ is the input at time-step $t$ in channel $n$. Each element describes how much an infinitesimal increase of input channel $n$ at time $t$ would raise ($G_c > 0$) or lower ($G_c < 0$) the class-$c$ readout at the end of the sequence, providing a two-dimensional visualization of the time-space regions that are either excitatory or inhibitory for a specific output neuron. The gradient is evaluated on an all-zero input, so the map reflects the network's intrinsic temporal structure rather than stimulus-specific effects. As in Methods~\ref{sec:methods_gradflow}, the readout is taken at the last time step of the sequence. The two analyses differ only in the scalar that is back-propagated and in the input.
\\

\noindent\textbf{Energy profiles.} To reduce each map to a scalar function of time, we define the gradient energy profile:
\begin{equation}
    E_c(t) \;=\; \sum_{n=1}^{N_{\text{in}}} G_c[t,n]^2 ,
\end{equation}
which aggregates the sensitivity of an output class to any perturbation of the input space, and is the quantity reported in Fig.~\ref{fig:gradmaps}e. Energies are reported in absolute value. The two models compared in this figure (learned and fixed recurrent delays) share the same architecture, training pipeline, and delay initialization, differing only in whether the delays were optimized. Their gradient magnitudes are therefore directly comparable, allowing both maps to be displayed on a common color scale (Fig.~\ref{fig:gradmaps}c,d).
\\

\noindent\textbf{Evaluation setting.} Models are rebuilt in the evaluation regime of Methods~\ref{sec:temporal benchmarks}: annealing spreads set to $\sigma = 0$ and delays rounded to integers, dropout disabled. We drive the networks with a zero input for T=3000 time steps, which is sufficient for the RSNNs to settle into their intrinsic dynamics. We then compute the gradient with respect to the final time step, and Fig.~\ref{fig:gradmaps} displays the last 150 time steps. Because spike thresholding is non-differentiable, the maps are computed through a surrogate gradient, for which we substitute a smooth arctangent \citep{fang2021incorporating} for the Triangle surrogate used during training, whose heavier tails widen the probed sensitivity and expose the longer-range temporal patterns of the receptive field. This is applied identically to all compared models.
\\

\noindent\textbf{Compared models.} We compute $G_c$ for each of the $35$ output classes, on four model classes sharing the SSC architecture of Methods~\ref{sec:temporal benchmarks}: an untrained network at its random initialization, a network with feedforward axonal delays trained as in \citep{Hammouamri2024}, and networks with learned and with fixed random axonal recurrent delays. Five independently seeded runs are analyzed per recurrent model class, and three for the feedforward-delay model. Learned and fixed delays models share a symmetric color scale, clipped at the $99$th percentile of $|G_c|$. The energy profiles of Fig.~\ref{fig:gradmaps}e are averaged over all output classes and seeds, and the shaded areas are the standard error of the mean over those $35 \times 5$ maps.

\subsection{Efficiency and robustness sweeps}
\label{sec:methods_constraints}

All sweeps in this section are run on HAR, with the pipeline, architecture and
hyperparameters of Methods~\ref{sec:temporal benchmarks}, over three seeds and for the four delay families (axonal or synaptic $\times$ learned or fixed). Unless stated otherwise, we report the final test accuracy of each run and read the delay statistics from the same checkpoint, in the inference regime (spread $\sigma = 0$, delays rounded to integers). Throughout, the "delay spread" of a network corresponds to the standard deviation of its recurrent delays, computed per layer and then averaged across seeds.
\\

\noindent\textbf{Delay initialization comparison (Fig.~\ref{fig:constraints}a).} Every family is trained at each value of the half-normal scale $s_{\text{init}} \in \{3, 8, 13, 18, 23\}$ used to draw the recurrent delays, $d^{\text{rec}}_i \sim |\mathcal{N}(0, s_{\text{init}}^2)|$. The lower panel reports the delay spread reached at the end of training. Both fixed-delay families achieve their highest accuracy at $s_{\text{init}} = 8$ ($82.7\%$ axonal, $82.9\%$ synaptic), which is therefore used in the two sweeps below.
\\

\noindent\textbf{Delay compression (Fig.~\ref{fig:constraints}b,c).} A network's ring buffer must hold as many time steps as its longest recurrent delay, and we study how short each model's delays can be made while still reaching a given accuracy. Learned delays are compressed with decoupled (AdamW) weight decay applied to the delay parameters alone, leaving the weight optimizer untouched, over $\gamma \in \{0, 0.001, 0.003, 0.01, 0.03, 0.04, 0.05, 0.06, 0.07, 0.1\}$. Fixed delays are never updated and cannot be compressed this way, so we instead shrink their initialization, $s_{\text{init}} \in \{0.5, 1, 2, 3, 5, 8\}$. Each value of $\gamma$ or $s_{\text{init}}$ contributes one (buffer depth, accuracy) point.
\\

\noindent\textbf{Firing-rate penalty (Fig.~\ref{fig:constraints}d,e).} To lower activity we add to the loss a penalty $\lambda\,C$ on the hidden spike trains as in \citep{Zimmer2019}:
\begin{equation}
    C \;=\; \frac{1}{L} \sum_{l=1}^{L} \frac{1}{2\,T\,N_l} \sum_{t=1}^{T}
    \sum_{n=1}^{N_l} \big(s^{\,l}[t,n]\big)^{2},
\end{equation}
where $s^{\,l}[t,n]$ is the spike emitted by neuron $n$ of hidden layer $l$ at time $t$, averaged over the batch. We sweep $\lambda \in \{0, 0.1, 0.3, 1, 3, 10, 30\}$ at $s_{\text{init}} = 8$ and without weight decay on the delays. Firing rate is measured on the test set as the mean number of spikes per hidden neuron and per time step, and each $\lambda$ contributes one (firing rate, accuracy) point.
\\

\noindent\textbf{Cost at matched accuracy (Fig.~\ref{fig:constraints}c,e).} Each value of $\gamma$ (or $s_{\text{init}}$ for the fixed-delays model) in \textbf{c}, and each value of $\lambda$ in \textbf{e}, yields, for a seed, a cost $c$ (the maximum recurrent delay, i.e., the buffer depth the model requires, in \textbf{c} and the mean firing rate in \textbf{e}) and a test accuracy $a$. Together, the configurations of one seed trace a cost-accuracy curve, but the panels \textbf{c} and \textbf{e} read that curve in the other direction: for each threshold $t$ on the grid $80, 80.5, \dots, 83\%$, they report the smallest cost at which the model attains $t$. We read this cost separately for each seed. Sorting that seed's configurations by increasing cost, $(c_1, a_1), (c_2, a_2), \dots$, let $i$ be the first index with $a_i \geq t$: the threshold is crossed between configurations $i-1$ and $i$, and we interpolate the cost linearly in accuracy between them,
\begin{equation}
    c(t) \;=\; (1 - w)\,c_{i-1} \;+\; w\,c_i, \qquad
    w \;=\; \frac{t - a_{i-1}}{a_i - a_{i-1}} \in [0, 1].
    \label{eq:iso_crossing}
\end{equation}
The plotted point is the mean of the three seed costs, the shaded band their SEM. A curve stops at the first threshold that one of its seeds never reaches. The full cost-accuracy frontiers, with every sampled configuration, are shown in Supplementary~\ref{supp:frontiers}.
\\

\noindent\textbf{Input perturbations (Fig.~\ref{fig:robustness}).} Robustness is measured on the networks of the initialization sweep, without retraining or any modification of the model: only the test inputs are perturbed, with three perturbations:
\begin{itemize}
    \item \textbf{temporal subsampling} by an integer factor $k \in \{1,2,3,4\}$: the samples at $t = 0, k, 2k, \dots$ are kept and the intermediate time steps set to zero.
    \item \textbf{deletion} with probability $p \in \{0, 0.1, \dots, 0.5\}$: each
    (time step, channel) input value is independently set to zero.
    \item \textbf{temporal jitter} of standard deviation $\sigma_{\text{jitter}} \in \{0, 1, 2, 4, 8, 12\}$ time steps: an offset $\delta_t \sim \mathcal{N}(0, \sigma_{\text{jitter}}^2)$ is drawn independently for every sample and time step and shared across input channels, then rounded to whole time steps and clamped at the window boundaries.
\end{itemize}
Deletion and jitter are stochastic, so each point is a mean across three model seeds
$\times$ three realizations. Subsampling is deterministic and is a mean across the three seeds alone.
\\

\noindent\textbf{Matched delay spread.} Training shifts the learned delays away from their initialization, so comparing the two conditions at the same $s_{\text{init}}$ would correspond to comparing different post-training delay distributions. Instead, we compare them at matched post-training spread. Let $a_c[m,k]$ be the accuracy of condition $c \in \{\text{learned}, \text{fixed}\}$ at perturbation magnitude $m$ and the $k$-th of the five initializations, averaged across seeds and realizations for stochastic perturbations, and let $s_c[k]$ be the corresponding delay spread. We construct a $60$-point grid spanning the range covered by both conditions, $\big[\max_c \min_k s_c[k],\ \min_c \max_k s_c[k]\big]$, so that nothing is extrapolated, and linearly interpolate $a_c[m,\cdot]$ onto it for each magnitude $m$
separately. The panel shows $\Delta = A_{\text{learned}} - A_{\text{fixed}}$ on that grid as a color map centered on zero, with the iso-accuracy contours of both conditions (learned solid, fixed dashed) and triangles marking each condition's five measured spreads (filled learned, open fixed).

\bmhead{Data availability}

All datasets used in this study are publicly available. The Spiking Speech
Commands (SSC) and Spiking Heidelberg Digits (SHD) datasets are available from
the Zenke lab at \url{https://zenkelab.org/datasets/} \citep{cramer2022heidelberg}. The permuted sequential MNIST (PS-MNIST) dataset is derived from MNIST as distributed with PyTorch/torchvision; the fixed pixel permutation used here is generated by the seeded script provided in the code repository. The Autonomous Localization (AL) and Human Activity Recognition
(HAR) datasets are those of the Neuromorphic Sequential Arena benchmark
\citep{chenNeuromorphicSequentialArena2025} available at \url{https://github.com/liyc5929/neuroseqbench}. The Mackey-Glass time
series are synthetic and are regenerated deterministically by the integration
script in the code repository, and the exact parameters are given in Methods. The data generated in this study, learned delay distributions, per-seed accuracies for every figure and table, gradient maps for all output classes and seeds, the profiling measurements and the trained models checkpoints are available at \url{https://github.com/alexmaxad/DelRec}. Finally, our results were produced using NVIDIA A100 GPUs and NVIDIA A40 GPUs for all datasets.

\bmhead{Code availability}

All code required to reproduce the results of this study, including the DelRec
layer implementations (PyTorch and fused Triton kernels), the training and
evaluation pipelines for all five datasets, and the analysis and plotting
scripts, is publicly available at \url{https://github.com/alexmaxad/DelRec}
under the MIT license.  The implementation builds on SpikingJelly \citep{spikingjelly}.

\bmhead{Competing interests}

The authors declare no competing interests.

\bmhead{Use of Large Language Models}

We declare the following use of LLMs in the preparation of this work:

\begin{itemize}
    \item For writing: OpenAI's ChatGPT and Anthropic's Claude were used in the writing of this article to improve the phrasing and the clarity of some sentences.
    \item For coding: OpenAI's ChatGPT and and Anthropic's Claude were used to enhance coding efficiency, more particularly for the factorization of existing scripts and the generation of plots.
\end{itemize}

\bmhead{Acknowledgements}

This work was supported by the French Defense Innovation Agency (AID) under grant number 2023 65 0082. We also thank Wei Fang and all the SpikingJelly developers for their excellent library \citep{spikingjelly}, and the quality of their support. This research was supported in part by the Agence Nationale de la Recherche under Grant ANR-20-CE45-0005 BRAIN-Net.

\putbib

\end{bibunit}

\newpage

\setcounter{page}{1}

\section{Supplementary}

\begin{bibunit}

\subsection{Difficulty and benefit of recurrent delays in SNNs simulators} \label{supp:Updating recurrent delays in SNNs simulators}

\begin{figure}
  \centering
  \includegraphics[width=1\linewidth]{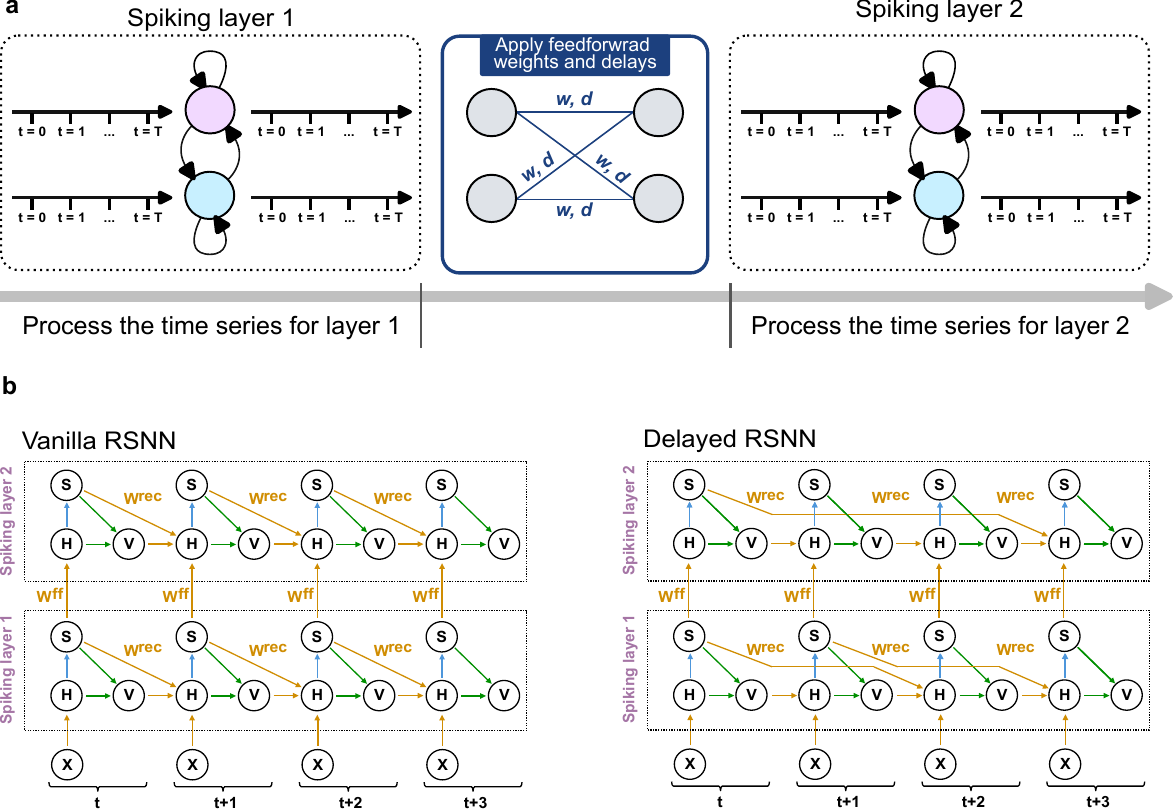}
  \caption{\textbf{Delays in recurrent connections: an implementation obstacle and a gradient advantage.} \textbf{a,} Time-series processing order in a RSNN with delays only in feedforward connections. Left: The input time-series are fully processed in the first spiking layer, resulting in a time-series of spikes per neuron. Center: Feedforward weights and delays between the first and second spiking layer can then be applied to all time-steps. Right: The time-series are fully processed by the second spiking layer. This is mathematically equivalent to a per time-step synchronous update of all layers. \textbf{b,} Computational graphs of a standard RSNN with an intrinsic delay of $1$ time step in all recurrent connections (Left), and of a RSNN with different and longer delays in the recurrent connections (Right). Variables $H, S$ and $V$ are defined in Eq.~\ref{eq neuronal charge} -~\ref{eq neuronal reset}. Delays in recurrent connections reduce the risks of exploding or vanishing gradients by bridging distant time steps.}
  \label{fig:hard_rec_delays}
\end{figure}

Extending delay learning to recurrent connections presents specific challenges that do not arise in the feedforward setting. Spiking Neural Networks simulators typically compute neuronal membrane dynamics in a layer-wise temporal loop and apply synaptic weights and delays only afterward, rather than updating the network in a fully synchronous manner. As depicted in Fig~\ref{fig:hard_rec_delays}, this update schedule makes it straightforward for methods such as DLCS-delays from \citep{Hammouamri2024} to incorporate and learn delays in feedforward networks, by inserting parallelizable computations between the temporal loops of successive spiking layers. Under this paradigm, extending delays to recurrent dynamics requires additional computations within the per-time-step update of each spiking layer, introducing challenges in efficiently simulating recurrent dynamics while handling the resulting increase in computational complexity.

By creating long gradient paths, recurrent connections make the gradients propagated by backpropagation through time (BPTT) particularly vulnerable to vanishing and exploding. However, in the presence of heterogeneous delays in the recurrent connections, a spike emitted at time $t$ can directly influence the network state at time $t + 1 + d_j$, bridging distant time steps without passing through all intermediate states, which can mitigate the risk of exploding (or vanishing) gradients due to the recurrent structure (see Supplementary~\ref{mitigate explo}).

\subsection{Datasets and hyperparameters} \label{sec:datasets}

\subsubsection{Datasets} 

\textbf{SSC}: The SSC dataset is a spiking audio benchmark that requires models to effectively exploit the temporal structure of spike trains to achieve high classification accuracy. While it
is one of the most widely used datasets in the SNN community for benchmarking
models' temporal processing capabilities, it also stands out as one of the largest,
featuring over $100$k samples across $35$ classes of spoken commands. It is worth
noting that this dataset has dedicated training, validation and test sets, and is
far from saturated (with best accuracies around $80\%$). We reduce the input
dimension by binning every $5$ of the original $700$ channels, yielding $140$ input
channels, and bin the spike trains temporally into $T = 250$ time steps of
$\Delta = 5.6$~ms.
\\ 

\noindent \textbf{PSMNIST}: The PS-MNIST is a vision dataset which is obtained by flattening all images of the MNIST ($28 \times 28$) into one sequence ($1 \times 784$), and permuting the pixel positions. This transformation disrupts local spatial structure and makes successful classification highly dependent on integrating long-range temporal dependencies, making PS-MNIST a standard benchmark for evaluating recurrent SNNs.
\\

\noindent \textbf{AL}: The AL dataset is a binary classification task in which a model must infer whether a robot can be found on the left or right plane after a sequence of actions (including ‘turn left’, ‘turn right’, ‘go straight’, and ‘stop’). Since the label depends on the whole action history, AL is well-suited to test the ability to model long-term dependencies. The dataset has 50k training samples for 5k test samples, all of 400 time-steps. As it lacks a dedicated validation set, we adopt the same evaluation procedure as \citep{chenNeuromorphicSequentialArena2025} to enable direct comparison with other models.
\\

\noindent \textbf{HAR}: The HAR dataset, requires classifying 18 activities from smartwatch gyroscope recordings from WISDM \citep{wisdm_smartphone_and_smartwatch_activity_and_biometrics_dataset__507}. Several classes differ only subtly (e.g., “eating chips” versus “eating pizza”), making the task particularly challenging and requiring the model to capture fine-grained short-term temporal patterns in noisy sensor signals. The dataset has 26048 training samples for 6512 test samples, all of 200 time-steps. It lacks a dedicated validation set, and we use the same procedure as \citep{chenNeuromorphicSequentialArena2025} to allow a fair comparison with the performance of other models.

\subsubsection{Hyperparameters} \label{hyperparameters}

We report here all hyperparameters used for the four temporal classification benchmarks. SSC and PS-MNIST follow the training recipe of \citep{xuASRCSNNAdaptiveSkip2025}, while AL and HAR are matched to the pipeline of \citep{chenNeuromorphicSequentialArena2025} at a $\approx 100$k-parameter budget. All networks use three fully connected hidden layers. On AL and HAR, however, the last hidden layer is a feedforward (non-recurrent) LIF layer, so only the first two hidden layers contain recurrent connections and learnable delays.

Feedforward weights and biases are initialized using the default PyTorch initialization for linear layers:
\[
w_{ij}, b_i \sim \mathcal{U}\!\left(-\sqrt{\tfrac{1}{\text{in\_features}}},\; +\sqrt{\tfrac{1}{\text{in\_features}}}\right).
\]
with $\text{in\_features}$ the number of input units of the layer. Recurrent weights are initialized orthogonally on SSC and PS-MNIST, and with the Kaiming-uniform scheme of \citep{chenNeuromorphicSequentialArena2025} on AL and HAR. Recurrent delays are drawn from a half-normal distribution $d^{\text{rec}}_i \sim |\mathcal{N}(0, s_{\text{init}}^2)|$ on SSC and HAR, and uniformly on $[0, 20]$ time-steps on PS-MNIST and AL (Table~\ref{tab:hyper-delay}). We chose the initial delay distributions which favored optimization stability. 

\begin{table}[htb]
\centering
\caption{General and optimization hyperparameters.}
\label{tab:hyper-general}
\renewcommand{\arraystretch}{1.15}
\small
\begin{tabularx}{\textwidth}{@{} X c c c c @{}}
\toprule
\textbf{Hyperparameter} & \textbf{SSC} & \textbf{PS-MNIST} & \textbf{AL} & \textbf{HAR} \\
\midrule
$\mathtt{hidden\ layers}$          & $[256,256,256]$ & $[64,212,212]$ & $[128,176,176]$ & $[128,176,176]$ \\
$\mathtt{recurrent\ hidden\ layers}$ & 3 & 3 & 2 & 2 \\
$\mathtt{epochs}$                  & 150 & 200 & 200 & 100 \\
$\mathtt{batch\ size}$             & 256 & 256 & 256 & 256 \\
$\mathtt{optimizer}$               & Adam & AdamW & AdamW & AdamW \\
$\mathtt{lr\ weights}$             & 1e-3 & 1e-3 & 5e-4 & 1.5e-3 \\
$\mathtt{lr\ delays}$              & 5e-2 & 1e-1 & 5e-3 & 5e-2 \\
$\mathtt{weight\ decay}$           & 1e-5 & 1e-2 & 0 & 0 \\
$\mathtt{scheduler\ weights}$      & One cycle & One cycle & Cosine & StepLR \\
$\mathtt{scheduler\ delays}$       & Cosine & Cosine & Cosine & StepLR \\
$\mathtt{gradient\ clipping}$      & --- & --- & 1.0 & --- \\
$\mathtt{feedforward\ dropout}$    & 0.1 & 0.1 & 0 & 0 \\
$\mathtt{recurrent\ dropout}$      & 0.3 & 0.2 & 0 & 0.2 \\
$\mathtt{batch\ norm}$             & $\mathtt{False}$ & $\mathtt{False}$ & $\mathtt{False}$ & $\mathtt{False}$ \\
$\mathtt{linear\ bias}$            & $\mathtt{True}$ & $\mathtt{True}$ & $\mathtt{True}$ & $\mathtt{True}$ \\
$\mathtt{recurrent\ bias}$         & $\mathtt{False}$ & $\mathtt{False}$ & $\mathtt{True}$ & $\mathtt{True}$ \\
\bottomrule
\end{tabularx}
{\footnotesize StepLR uses $\mathtt{step\ size}=10$, $\gamma=0.8$. The delays are optimized by a second optimizer (``delays'') with its own learning rate and no weight decay.}
\end{table}

\begin{table}[htb]
\centering
\caption{Neuron hyperparameters.}
\label{tab:hyper-neuron}
\renewcommand{\arraystretch}{1.15}
\small
\begin{tabularx}{\textwidth}{@{} X c c c c @{}}
\toprule
\textbf{Hyperparameter} & \textbf{SSC} & \textbf{PS-MNIST} & \textbf{AL} & \textbf{HAR} \\
\midrule
$\tau \; (\text{time-steps})$   & 2 & 2 & 2 & $1/0.9$ \\
$V_{threshold}$                 & 1.0 & 1.0 & 0.5 & 0.5 \\
$\mathtt{reset\ type}$          & Soft & Soft & Hard & Hard \\
$V_{reset}$                     & --- & --- & 0 & 0 \\
$\mathtt{detach\ reset}$        & $\mathtt{False}$ & $\mathtt{False}$ & $\mathtt{False}$ & $\mathtt{False}$ \\
surrogate function              & Triangle & Triangle & Triangle & Triangle \\
surrogate width $\alpha$        & 1.0 & 1.0 & 0.4 & 0.6 \\
\bottomrule
\end{tabularx}
\end{table}

\begin{table}[htb]
\centering
\caption{Delay-learning hyperparameters. HAR scores are reported for the best delay-distribution width from the sweep of Section~\ref{sec:constraints}, so its delay-init width is swept rather than fixed.}
\label{tab:hyper-delay}
\renewcommand{\arraystretch}{1.15}
\small
\begin{tabularx}{\textwidth}{@{} X c c c c @{}}
\toprule
\textbf{Hyperparameter} & \textbf{SSC} & \textbf{PS-MNIST} & \textbf{AL} & \textbf{HAR} \\
\midrule
delay initialization            & Half-normal & Uniform & Uniform & Half-normal \\
init parameter               & $s_{\text{init}}=12$ & $[0,20]$ & $[0,20]$ & $s_{\text{init}} \in \{3,\dots,23\}$ \\
$\sigma_{init}$                 & 1.0 & 1.0 & 0 & 0 \\
$\sigma_{decay}$                & 0.95 & 0.8 & 0.95 & 0.95 \\
round delays at inference       & $\mathtt{True}$ & $\mathtt{True}$ & $\mathtt{False}$ & $\mathtt{True}$ \\
recurrent weight init           & Orthogonal & Orthogonal & Kaiming-unif. & Kaiming-unif. \\
\bottomrule
\end{tabularx}
\end{table}

\subsubsection{Statistical significance} \label{sup:significance}

We assess the significance of the differences reported in Table~\ref{tab:ablation} using a two-sided Welch’s t-test for unequal variances, which assumes only that the two sets of runs are independent. All eight learned-versus-fixed differences favor learned delays, with six reaching statistical significance at $p < 0.05$ (Table~\ref{tab:signif-ablation}). The two exceptions are the synaptic configurations on SSC ($p = 0.28$) and AL ($p = 0.10$), the two datasets on which
synaptic delays bring no accuracy benefit (Section~\ref{sec:tasks}). The same test applied to the learned axonal-synaptic pairs confirms that neither delay type
dominates: axonal is ahead on SSC ($+0.18$ pts, $p = 0.21$) and AL ($+0.74$, $p = 0.37$), and behind on PS-MNIST ($-0.31$, $p = 0.020$) and HAR ($-0.91$, $p = 0.013$). Finally, most baselines of Table~\ref{tab:SSC-PSMNIST} and Table~\ref{tab:AL-HAR} are single runs and cannot be
tested, but among those reporting several seeds, DelRec exceeds SiLIF~\citep{fabreStructuredStateSpace2025} by $0.39$ pts ($p = 0.014$) and DCLS~\citep{Hammouamri2024} by $1.73$ pts ($p < 10^{-5}$) on SSC.

\begin{table}
\centering
\caption{\textbf{Learned versus fixed recurrent delays.} Difference in mean accuracy
(learned $-$ fixed), in accuracy points, and the two-sided Welch $t$-test $p$-value.
$n = 5$ seeds, except HAR ($n = 3$).}
\label{tab:signif-ablation}
\renewcommand{\arraystretch}{1.1}
\setlength{\tabcolsep}{4pt}
\footnotesize
\begin{tabularx}{\textwidth}{@{} X l c c @{}}
\toprule
Dataset & Delay type & $\Delta$ acc.\ [pts] & $p$ \\
\midrule
SSC       & Axonal   & $+0.31$ & $0.041$  \\
SSC       & Synaptic & $+0.15$ & $0.275$  \\
PS-MNIST  & Axonal   & $+0.55$ & $0.0073$ \\
PS-MNIST  & Synaptic & $+0.73$ & $0.0005$ \\
AL        & Axonal   & $+3.02$ & $0.016$  \\
AL        & Synaptic & $+1.27$ & $0.101$  \\
HAR       & Axonal   & $+0.79$ & $0.045$  \\
HAR       & Synaptic & $+1.55$ & $0.0021$ \\
\bottomrule
\end{tabularx}
\end{table}

\subsection{Additional results}

\subsubsection{NMSE values across all $\tau$ and $H$ for the Mackey-Glass prediction task}\label{sup:mg-bars}

\begin{figure}
  \centering
  \includegraphics[width=\linewidth]{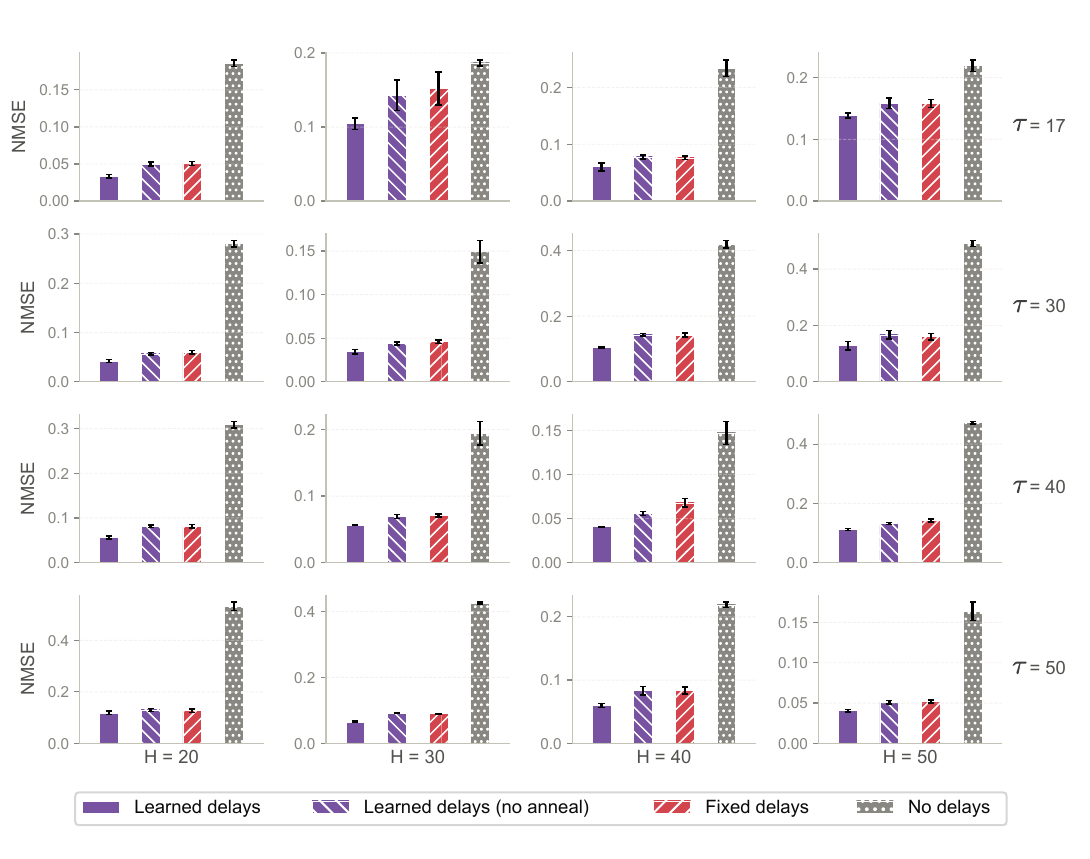}
  \caption{\textbf{Absolute test NMSE on the Mackey-Glass task for all $(\tau, H)$ conditions.} Each panel is one $(\tau, H)$ condition, with the chaos parameter $\tau$ across rows and the prediction horizon $H$ across columns, and compares the four models: the full DelRec model (learned delays with $\sigma$-annealing), the same model without annealing, fixed random recurrent delays and an RSNN without delays. Bars are the mean over 3 seeds and error bars their SEM.}
  \label{fig:sup-mg-bars}
\end{figure}

Fig.~\ref{fig:mackey-glass}a-c reports the relative NMSE reduction of DelRec over each baseline. Fig.~\ref{fig:sup-mg-bars} gives the absolute test NMSE behind those numbers, for the four variants and all $(\tau, H)$ conditions. The full DelRec model has the lowest error in every condition. Absolute values are not directly comparable across panels, as an independent time series is generated for each $\tau$, so the difficulty of a given horizon $H$ varies with the attractor.

\subsubsection{Comparison with a single delay per layer} \label{sec:single delay}

Here, we compare our delay learning method with the architecture used in \citep{xuASRCSNNAdaptiveSkip2025} using a single delay parameter per layer (obtained by a different algorithm). We implemented this shared
("common") delay within DelRec, changing only the shape of the delay parameter: each
recurrent layer $L$ now carries a single scalar $d^{(L)}$, broadcast to all of its
recurrent connections in Eq.~\ref{eq:rec_update}, that is $2$ delay parameters on HAR instead of $304$ (axonal) or $47{,}360$ (synaptic). The architecture, training pipeline, hyperparameters, and random seeds are identical to those of the four heterogeneous families. The delays $d^{(L)}$ are initialized from the same half-normal distribution $|\mathcal{N}(0, s_{\mathrm{init}}^2)|$, and optimized using the same surrogate-gradient rule with the same learning rate (Methods~\ref{sec:learning_procedure}). We sweep $s_{\mathrm{init}}$ 
 over the same grid, using three seeds per configuration.

\begin{figure}
  \centering
  \includegraphics[width=\textwidth]{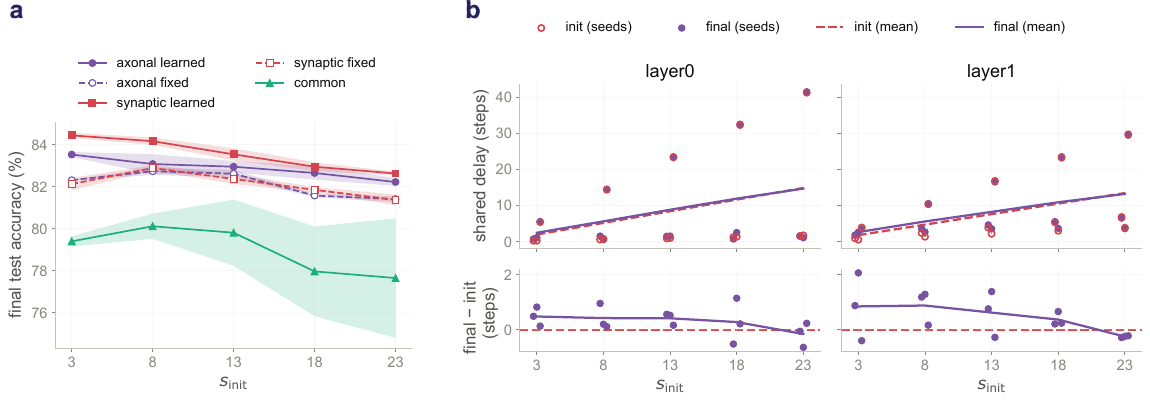}
  \caption{\textbf{Performance and evolution of a single delay per layer.} HAR, sweep over the scale $s_{\mathrm{init}}$ of the half-normal delay
  initialization, three seeds per point.
  \textbf{a}, Final test accuracy against $s_{\mathrm{init}}$ for the four heterogeneous configurations and for the shared-delay variant, mean $\pm$ SEM over seeds (shaded). \textbf{b}, Evolution of the shared delay itself, one column per recurrent layer. Top: value of the shared delay at initialization (open red circles) and after training (filled purple circles, purple solid line). Bottom: the displacement (final $-$ initial) on its own scale, against the zero line.}
  \label{fig:commondelay}
\end{figure}

The shared-delay model never reaches the heterogeneous configurations
(Fig.~\ref{fig:commondelay}a), and, at every $s_{\mathrm{init}}$, is below the fixed families, which involve no delay optimization at all. Its dispersion across seeds also grows with $s_{\mathrm{init}}$, presumably because of the difficulty for the model to handle a large single delay draw. Fig.~\ref{fig:commondelay}b shows that the shared delay barely moves during training, so every run ends close to where its initialization put it,
and its accuracy is largely decided by a single random draw. We hypothesize that this lack of evolution results from the effect of updating a shared parameter: a single delay update is more costly than in the heterogeneous case because a shared delay re-times the whole layer at once, so any change simultaneously perturbs every recurrent input the weights have adapted to, whereas a heterogeneous configuration can be reorganized neuron by neuron. We also hypothesize that a shared parameter prevents the layer from receiving a clear learning signal, as it accumulates delay gradients from all recurrent connections, whose potentially conflicting contributions may cancel each other out. What delay learning exploits is thus not only the presence of a tunable lag but the heterogeneity of the delay configuration, which justifies parameterizing delays per neuron or per connection despite the additional parameters.

\subsubsection{Perturbations} \label{sup:perturbations}

\begin{figure}
  \centering
  \includegraphics[width=\textwidth]{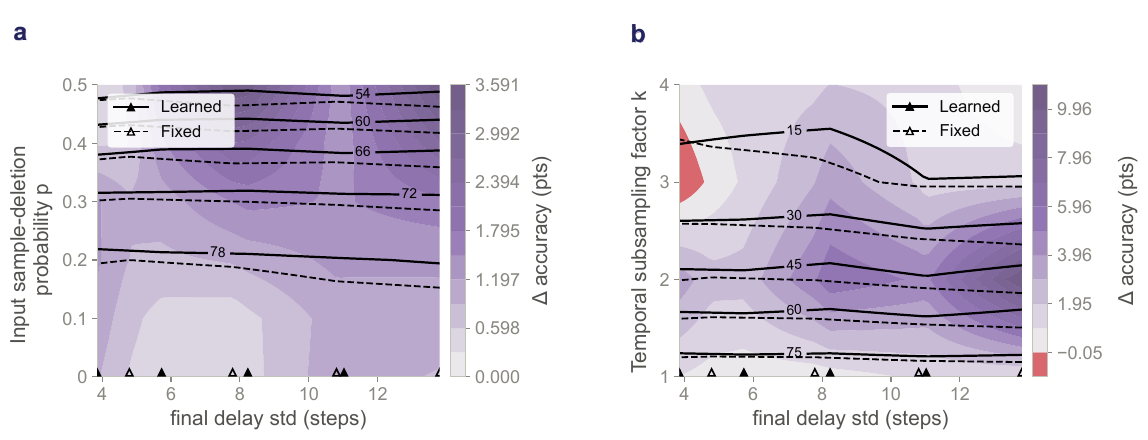}
  \caption{\textbf{Learned axonal delays are also more robust to input deletion and
  subsampling.} Accuracy difference between learned and fixed axonal recurrent delays ($\Delta$ accuracy $=$ learned $-$ fixed) as a function of the perturbation strength (y-axis) and of the final recurrent-delay spread (x-axis), as in Fig.~\ref{fig:robustness}. \textbf{a}, Random input-sample deletion with probability $p$. \textbf{b}, Temporal subsampling by factor $k$. Solid and dashed lines are the iso-accuracy contours of the learned and fixed models respectively, and filled and open triangles mark the final delay std values they reach. Models come from the initialization sweep ($s_{\mathrm{init}} \in \{3, 8, 13, 18, 23\}$). Mean over 3 model seeds $\times$ 3 perturbation realizations for \textbf{a}, and over 3 model seeds for \textbf{b}.}
  \label{fig:axonal_pert}
\end{figure}

\noindent\textbf{Axonal counterparts of Fig.~\ref{fig:robustness}a,b.}
Fig.~\ref{fig:robustness}a,b report the subsampling and deletion sweeps for synaptic delays. Fig.~\ref{fig:axonal_pert} shows the axonal ones, obtained with the identical protocol (Methods~\ref{sec:methods_constraints}). The two delay types behave the same way: the accuracy difference between learned and fixed delays is positive over essentially the whole plane spanned by the perturbation magnitude and the final delay spread, the only exception being a small patch at the shortest spread under strong subsampling. The advantage of delay learning under these two
perturbations is therefore a property of recurrent delay learning rather than of a particular delay parameterization, unlike the response to temporal jitter
(Fig.~\ref{fig:robustness}c,d), which separates the two types.

\subsubsection{Frontiers under constraints}

\begin{figure}
  \centering
  \includegraphics[width=0.7\textwidth]{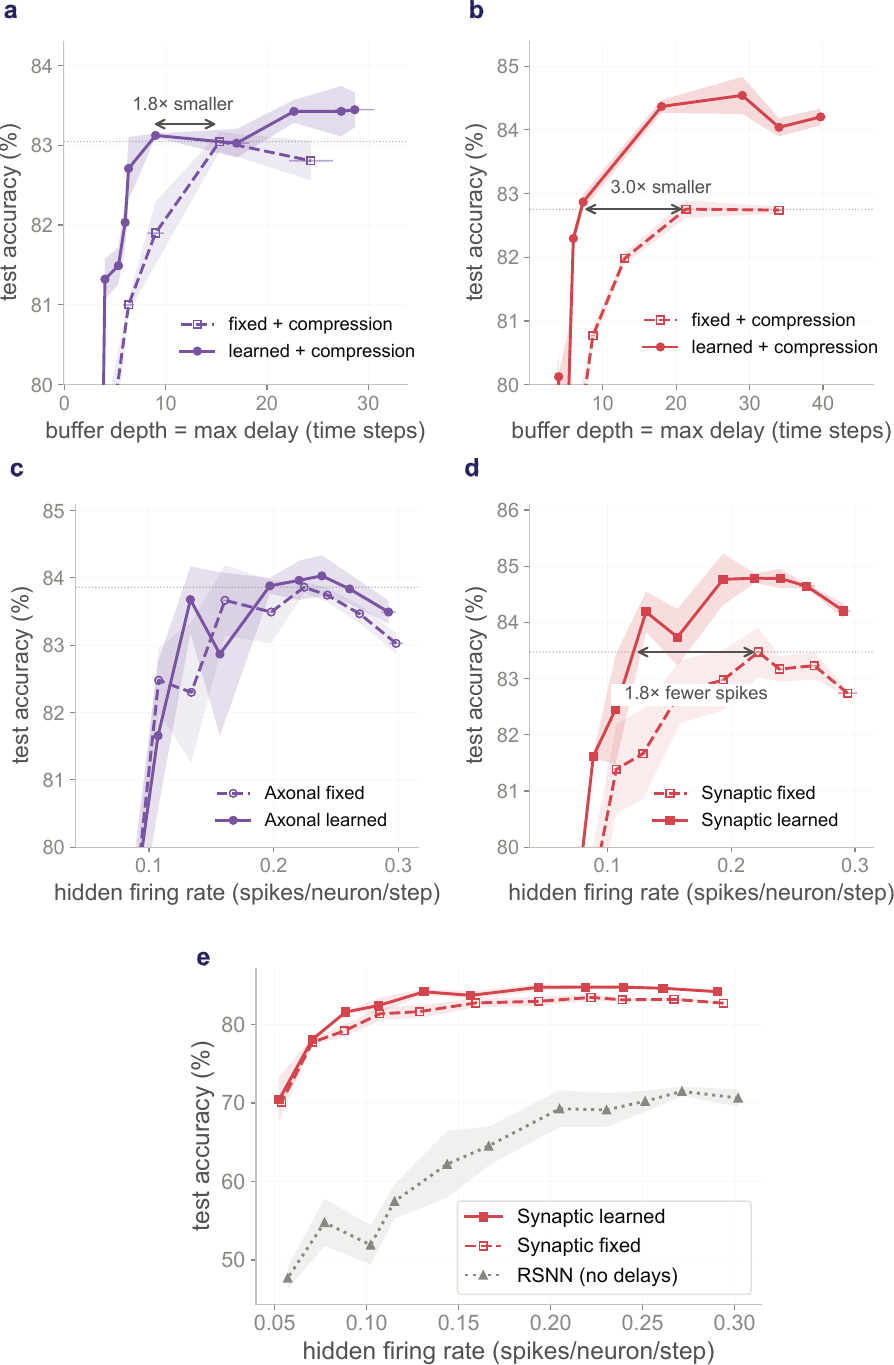}
  \caption{\textbf{Cost-accuracy frontiers on HAR: the curves behind
  Fig.~\ref{fig:constraints}c,e.} Three seeds per configuration, each point is one configuration, at the mean over its seeds of the cost and of the test accuracy (shaded band is $\pm$SEM on accuracy, whisker is $\pm$SEM on cost). Dotted lines are the best accuracy the corresponding fixed model attains at any budget. Double arrows (\textbf{a}, \textbf{b},
  \textbf{d}) are the iso-accuracy cost ratios at that level, interpolated as in
  Fig.~\ref{fig:constraints}c,e. \textbf{a}, \textbf{b}, Accuracy against buffer depth for axonal (\textbf{a}) and synaptic (\textbf{b}) delays. Learned delays are compressed by weight decay on the delay parameters, fixed delays by' a tighter initialization
  (Methods~\ref{sec:methods_constraints}). \textbf{c}, \textbf{d}, Accuracy against the hidden firing rate over the spike-penalty sweep
  ($\lambda = 0$ at the right) for axonal (\textbf{c}) and synaptic
  (\textbf{d}) delays. \textbf{e}, The same synaptic sweep against an RSNN without delays, trained on the same grid of $\lambda$, shown on the full accuracy range. \textbf{a}--\textbf{d} are cropped below $80\%$.}
  \label{supp:frontiers}
\end{figure}

Fig.~S\ref{supp:frontiers} shows the frontiers that Fig.~\ref{fig:constraints}c,e read at fixed accuracy (Methods~\ref{sec:methods_constraints}). For buffer depth, (Fig.~S\ref{supp:frontiers}a,b), the learned frontiers lie
above their fixed controls over the whole overlapping range. For firing rate
(Fig.~S\ref{supp:frontiers}c,d), learned synaptic delays lie above their control at every $\lambda$ and reach, whereas the two axonal curves are indistinguishable, crossing repeatedly with differences that never exceed the spread across seeds, the asymmetry reported in Fig.~\ref{fig:constraints}e. Fig.~S\ref{supp:frontiers}e places the same synaptic sweep against an RSNN without delays, trained
on the same grid, which never reaches the threshold range of
Fig.~\ref{fig:constraints}e. Both delay families exceed the best accuracy the RSNN attains at any budget ($71.5\%$) while spending less than a third of its spikes, and the gap widens as activity is removed.

\subsubsection{Effect of delay rounding and alternatives} \label{delay rounding effect}

\begin{figure}
  \centering
  \includegraphics[width=\textwidth]{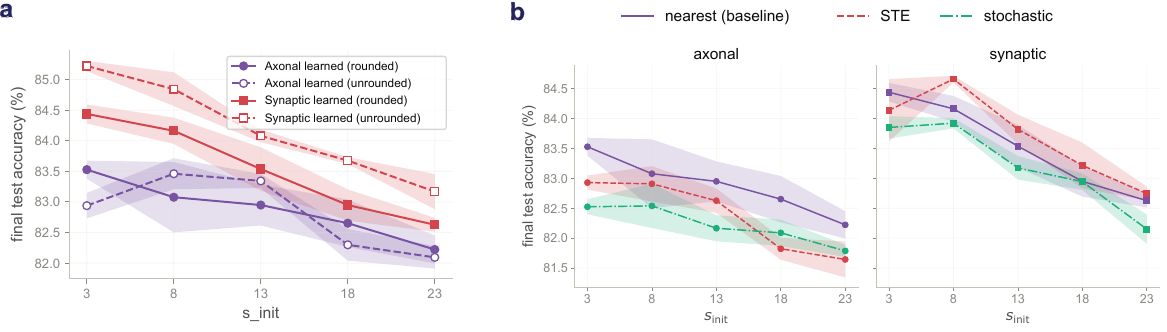}
  \caption{\textbf{On HAR, delay discretization has a limited cost and no alternative improves on it.} Sweep over the scale $s_{\mathrm{init}}$ of the half-normal delay initialization, three seeds per point, mean $\pm$ SEM as shaded areas. \textbf{a}, Learned axonal and synaptic delays, trained and tested rounded to the nearest integer at every epoch (solid, filled markers) or kept fractional (dashed, open markers). \textbf{b}, Axonal (left) and synaptic (right) learned delays under the three discretizations: nearest rounding (the default), the straight-through estimator and stochastic rounding.}
  \label{fig:rounding}
\end{figure}

Once $\sigma$ has been annealed to zero, a fractional delay is a linear interpolation between the two closest integer lags and occupies two taps of the scheduling buffer instead of one. Since integer delays are what neuromorphic hardware implements, we project the delays back onto the integer grid at every epoch and at test time ($d \leftarrow \lfloor d \rceil$). On HAR, we compare this projection with training without it (i.e., keeping delays fractional throughout training) and with two alternative discretization schemes: a straight-through estimator (STE), which takes $\text{int}(d)$ in the forward pass, while the gradient flows through to the fractional relaxation in the backward pass, leaving no train-test discretization gap. We also compare with stochastic rounding, $d \leftarrow \lfloor d + \mathcal{U}[0,1) \rfloor$. The projection has a very limited cost on HAR (Fig.~\ref{fig:rounding}a). Axonal delays are indistinguishable in the two regimes, while synaptic delays lose a consistent $0.5$ to $0.8$ points to the fractional runs. Neither alternative (straight through or stochastic rounding) improves on nearest rounding (Fig.~\ref{fig:rounding}b). For simplicity, we stuck to delay rounding at each epoch, which we believe also serves as a regularizer.

\begin{figure}
  \centering
  \includegraphics[width=0.6\textwidth]{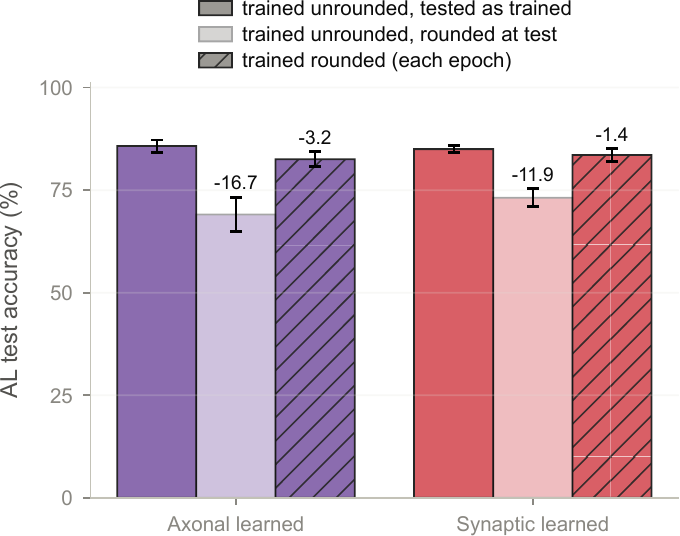}
  \caption{\textbf{Cost of discretization on the AL dataset.} AL test accuracy of learned axonal and synaptic delays under three regimes: delays kept fractional during training and at test, the same models rounded to the nearest integer at test only, and models rounded at every epoch during training and at test. Numbers above the bars are the accuracy drop relative to the first regime. Mean $\pm$ SEM over 5 seeds.}
  \label{fig:al_discretization}
\end{figure}

On AL, in contrast, the same projection is costly (Fig.~\ref{fig:al_discretization}), which is why the AL scores of Table~\ref{tab:ablation} are reported with $\sigma = 0$ and unrounded delays, ie. with each delay represented as a weighted combination of its two nearest integers (Section~\ref{sec:learning-delays}). This degradation is however a train-test mismatch rather than an intrinsic cost of integer delays: rounding at every epoch during training, as we do on the three other benchmarks, brings the loss down to $3.2$ and $1.4$ points respectively. Yet, even in that fully integer, hardware-deployable regime the AL accuracy remains above the best published baseline on that dataset. Integer delays therefore remain available on AL at a limited cost, and all our models stay deployable on hardware with programmable integer delays.

\subsection{Computational overhead} \label{computational_overhead}

The DelRec method enables state-of-the-art performances in dynamic processing with very competitive parameter counts, but it comes at the expense of buffer computations nested in the temporal loop of the recurrent spiking neurons (see Algorithm~\ref{alg:learn-delays}). While regular RSNN layers have a temporal complexity of $O(BS)$ with $S = N_{in} N_{out}$ the number of synapses in the recurrent connections, DelRec increases the complexity to $O(BSL)$ where $L$ is the effective length of the delay buffer (which is of the order of $\max_i d_i$ with $i$ the neurons in the layer). The same logic applies to memory overhead. In PyTorch, automatic differentiation exacerbates this issue: each write to the scheduling buffer creates a node in the computational graph, forcing a separate copy of the buffer to be retained at every time step until the backward pass. As a result, the memory footprint of a delayed layer grows with both the maximum delay and the delay spread.

We therefore implemented the delayed recurrent layer as fused GPU kernels written in Triton, one for each delay type, which keep the whole temporal loop within a single kernel launch and recompute the scheduling buffer in the backward pass instead of storing it. The two delay types call for different strategies, which are detailed in the public repository \url{https://github.com/alexmaxad/DelRec}. Both kernels are exact, and reproduce the forward values as well as the gradients with respect to weights and delays of the PyTorch implementation up to numerical precision.

\begin{figure}
  \centering
  \includegraphics[width=\textwidth]{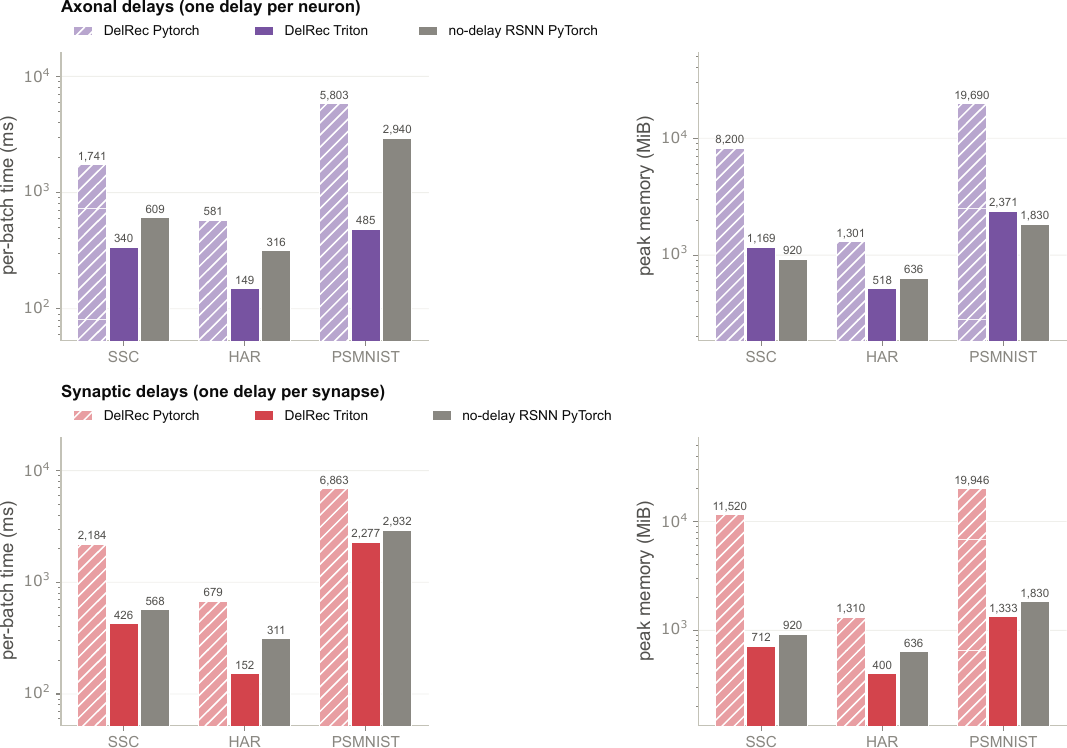}
  \caption{\textbf{Triton kernels bring the cost of a delayed recurrent layer down to the one of a conventional RSNN.} Per-batch time (left) and peak GPU memory (right) of a full forward and backward pass on real batches, for axonal (top) and synaptic (bottom) recurrent delays, on the SSC, HAR and PS-MNIST datasets. Each dataset is profiled at its own operating point, i.e., with the sequence length, batch size and layer widths of its configuration (Supplementary~\ref{hyperparameters}). Bars compare the PyTorch implementation of Algorithm~\ref{alg:learn-delays} (hatched), the corresponding Triton kernel (solid), and a PyTorch RSNN without delays sharing the same architecture (grey). Median over $3$ runs of $4$ batches on a NVIDIA A40.}
  \label{fig:kernel_datasets}
\end{figure}

\begin{figure}
  \centering
  \includegraphics[width=\textwidth]{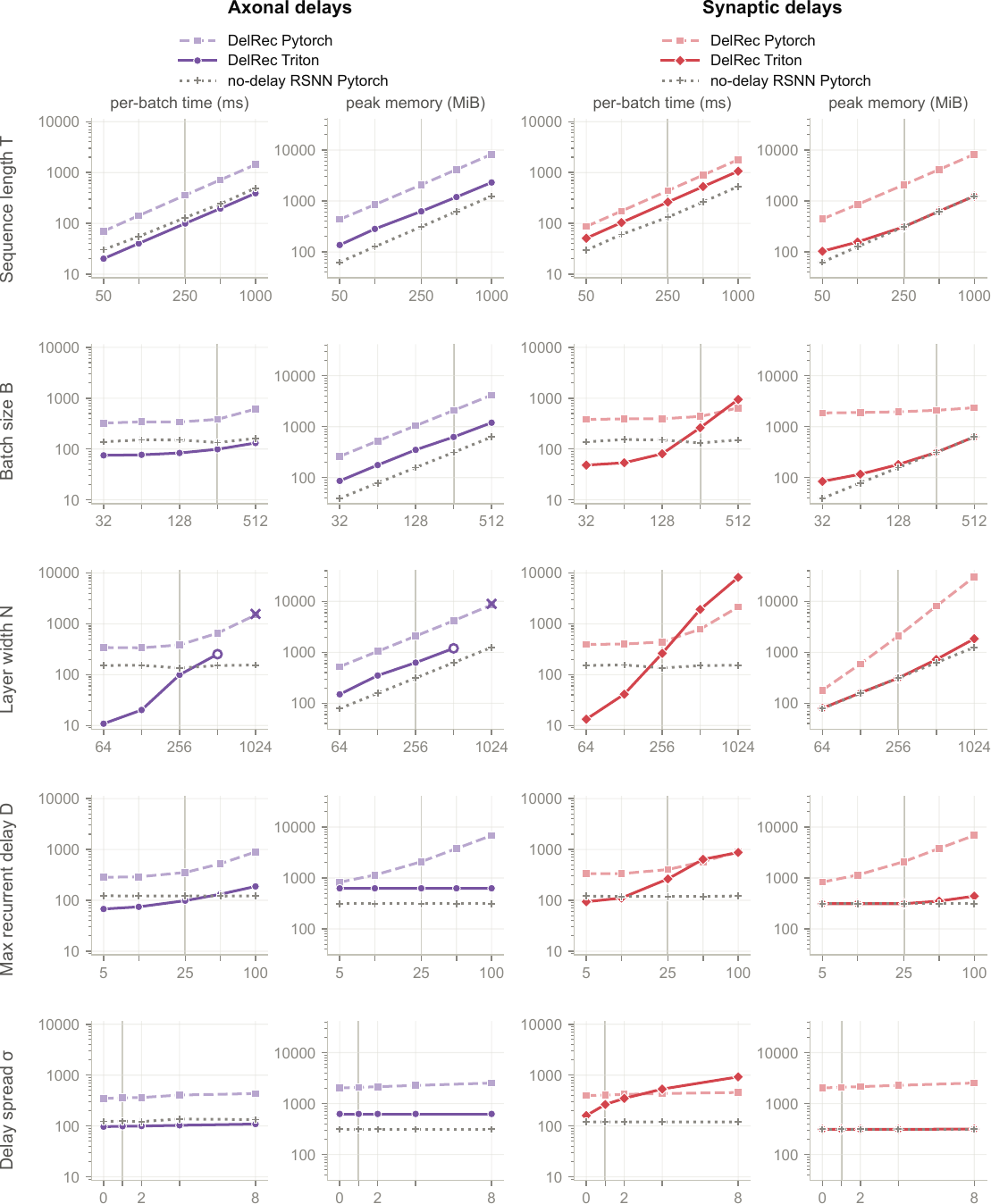}
  \caption{\textbf{Scaling of a delayed recurrent layer along five axes.} Per-batch time and peak GPU memory of a single recurrent layer (forward and backward), for axonal (left) and synaptic (right) delays, sweeping one factor at a time around $T = 250$, $B = 256$, $N = 256$, $D = 25$, $\sigma = 1$ (grey vertical line): the sequence length $T$, the batch size $B$, the layer width $N$, the maximum recurrent delay $D$ and the delay spread $\sigma$. Median over $10$ runs on a NVIDIA A40. Markers sit on all five swept values. Empty markers denote a narrower weight tile ($BK = 32$) in the Triton parallelization, the default $BK = 64$ exceeding the $99$ KiB of shared memory of the A40, and $\times$ denotes that no tile fits at all.}
  \label{fig:kernel_sweeps}
\end{figure}

We performed two profiling experiments in order to characterize the resulting computational cost, both on a NVIDIA A40 GPU, measuring a full forward and backward pass. In the first one (Fig.~\ref{fig:kernel_datasets}), we profile the complete model on real batches, for each dataset with the sequence length, batch size and layer widths of its configuration (Supplementary~\ref{hyperparameters}), and we compare the PyTorch implementation of Algorithm~\ref{alg:learn-delays}, the corresponding Triton kernel, and a standard RSNN without delays sharing the same architecture. In the second one (Fig.~\ref{fig:kernel_sweeps}), we profile a single recurrent layer and sweep one factor at a time around a reference configuration ($T = 250$, $B = 256$, $N = 256$, $D = 25$, $\sigma = 1$), in order to isolate the contribution of each of them. Additional hyperparameters are fixed and shared across models, and are available in the corresponding script of \url{https://github.com/alexmaxad/DelRec}.

The Triton kernels considerably reduce the per-batch time and the peak memory over the PyTorch implementation (Fig.~\ref{fig:kernel_datasets}). On a GPU, a delayed layer is thereby no longer more expensive than a conventional one used with plain Pytorch, and in all six configurations, the delayed model trains faster for a comparable memory.

The sweeps in Fig.~\ref{fig:kernel_sweeps} reveal where the remaining cost lies. The maximum delay and the spread, which dominated the PyTorch implementation, no longer affect memory, and only the layer width remains unfavorable, so delays are best suited to small and medium-sized layers, which is also the regime in which our models operate (Supplementary~\ref{hyperparameters}). Importantly, these measurements correspond to training costs. At inference, delays are rounded to integers, $\sigma$ is null and no intermediate states need to be retained for backpropagation. The overhead therefore reduces to the intrinsic hardware cost of implementing delays on-chip (see \citep{meszaros2025completepipelinedeployingsnns} for a detailed study of these costs). Consequently, the temporal expressivity afforded by recurrent delays is achieved at a computational cost comparable to that of a conventional RSNN, both during training and deployment.

\clearpage
\newpage

\subsection{The case of the SHD dataset}

\subsubsection{SHD dataset} \label{SHD appendix}

The SHD (Spiking Heidelberg Digits) dataset \citep{cramer2022heidelberg} is currently one of the most widely used dataset in the SNN community to benchmark spatiotemporal processing. It is a much smaller dataset than SSC or PS-MNIST, with $10$k recordings of spoken digits ranging from zero to nine, in English and German. While the SHD dataset has historically been a valuable tool to help SNNs evolve towards better classification performances, we argue that this dataset has reached saturation and became obsolete in terms of benchmarking for the following two reasons:
\begin{itemize}
    \item SHD lacks a dedicated validation set, and historical evaluations of SNN performance on this dataset have relied solely on the test set. This approach is methodologically flawed and leads to an overfitting of the test set. More recent works have set more rigorous standards by using a fraction of the training set as a validation set, before reporting the best model's accuracy on the test set ~\citep{Baronig2025, meszaros2025efficienteventbaseddelaylearning}.
    \item While the best models report around $93\%$ of accuracy on the test set (using a clean split), ~\citep{meszaros2025efficienteventbaseddelaylearning} explain that further improvements in performance are likely not statistically significant given the small size of the test set ($2264$): with naive assumptions on error rates, the Bayesian confidence intervals of accuracies over $93\%$ overlap.
\end{itemize}
For all these reasons, we decided not to include SHD in Table~\ref{tab:SSC-PSMNIST}, and we recommend its use only as an initial validation step for proof-of-concept studies. Nevertheless, we evaluated DelRec on the SHD dataset, and we provide in this Appendix the results and the methodology used for these experiments. 

\subsubsection{Methodology}

For this dataset, we also reduced the input size by binning every $5$ neurons of the original $700$, resulting in a spatial input dimension of $140$. We also binned the inputs temporally, using a discrete time-step of $\Delta = 10\text{ms}$. The architecture used is composed of two fully connected hidden layers, and each of these layers can be followed by batch normalization, a neuron module, and dropout. The readout consists of $n_{\text{classes}} = 20$ leaky-integrate neurons with an infinite threshold, so that they never spike and their membrane potential integrates the activity of the last hidden layer. Following \citep{Bittar2022}, we take the
softmax over the readout neurons at each time step and average it over time. Denoting $V^{r}[t]\in\mathbb{R}^{n_{\text{classes}}}$ the vector of readout membrane potentials at time step $t$, and $T$ the number of time steps, the predicted class distribution for one sample is
\begin{equation}
  \hat{y}_i \;=\; \frac{1}{T}\sum_{t=1}^{T}
  \big[\operatorname{softmax}\big(V^{r}[t]\big)\big]_i.
\end{equation}
Denoting $N$ the batch size,
$\hat{y}^{(n)}$ the distribution obtained for sample $n$ and $y_n$ its label, the
loss for one batch is the cross-entropy
\begin{equation}
  \mathcal{L} \;=\; -\frac{1}{N}\sum_{n=1}^{N} \log \hat{y}^{(n)}_{y_n}.
\end{equation}

We used augmentations on the training set of the SHD in order to limit overfitting. We use this method only on our large models when trying to reach high accuracies, not in our ablation study. We followed the approach of \citep{Nowotny_2025} and \citep{meszaros2025efficienteventbaseddelaylearning}:

\begin{itemize}
    \item We applied a random temporal shifting to inputs, with a shift uniformly drawn in the interval $\big[-100 \text{\ time\ steps} , \; 100 \text{\ time\ steps} \big]$.
    \item Each input was blended with an input from the same class by aligning their center of mass and, for each time step, picking a spike from either sample with a probability of $0.5$.
\end{itemize}

We initialized the feedforward weights using Kaiming uniform initialization, with $a = \sqrt{5}$: 
\[
w_{ij} \sim \mathcal{U}\!\left(-\sqrt{\tfrac{6}{\text{in\_features}}},\; +\sqrt{\tfrac{6}{\text{in\_features}}}\right),
\]
\[
b_i \sim \mathcal{U}\!\left(-\tfrac{1}{\sqrt{\text{in\_features}}},\; +\tfrac{1}{\sqrt{\text{in\_features}}}\right),
\]
where \(\text{in\_features}\) denotes the number of input units to the layer.
The recurrent delays were initialized with a random draw from a uniform distribution:  
\[
d_{i}^{\text{rec}} \sim \mathcal{U}(10,\;30).
\]

The hyperparameters used are presented in Table~\ref{tab:shd-hyper} and Table~\ref{tab:shd_neurons-hyper}.

\begin{table}
\centering
\caption{General hyperparameters used for the SHD dataset.}
\label{tab:shd-hyper}
\renewcommand{\arraystretch}{1.15} 
\small
\begin{tabularx}{\textwidth}{@{} X c @{}}
\toprule
\textbf{Hyperparameter} &  \textbf{SHD} \\
\midrule
$\mathtt{epochs}$          & 150 \\
$\mathtt{batch\ size}$      & 256 \\
$\mathtt{learning\ rate\ max\ weights}$   & 5e-3 \\
$\mathtt{learning\ rate\ max\ positions}$   & 5e-2 \\
$\mathtt{optimizer}$        & AdamW \\
$\mathtt{weight\ decay}$    & 1e-5 \\
$\mathtt{batch\ norm}$      & $\mathtt{True}$ \\
$\mathtt{bias}$             & $\mathtt{True}$ \\
$\mathtt{feedforward\ dropout}$           & 0.4 \\
$\mathtt{recurrent\ dropout}$             & 0.2 \\
$\mathtt{scheduler\ weights}$             & One cycle \\
$\mathtt{scheduler\ positions}$           & Cosine annealing \\
\bottomrule
\end{tabularx}
\end{table}

\begin{table}
\centering
\caption{Neuron-related hyperparameters used for the SHD dataset.}
\label{tab:shd_neurons-hyper}
\renewcommand{\arraystretch}{1.15} 
\small
\begin{tabularx}{\textwidth}{@{} X c @{}}
\toprule
\textbf{Hyperparameter}  & \textbf{SHD} \\
\midrule
$\mathtt{\tau} \; (\text{time-steps})$  & 1.005 \\
$V_{threshold}$                  & 1. \\
$\mathtt{reset\ type}$           & Hard \\
$V_{reset}$                      & 0 \\
$\mathtt{detach\ reset}$         & $\mathtt{True}$ \\
surrogate function               & Arctan \\
\bottomrule
\end{tabularx}
\end{table}

\subsubsection{Results}

We present the performance of DelRec on the SHD dataset in Table~\ref{tab:shd-perf-table}. The model reaches competitive performance, with a significantly low number of parameters. 

\begin{table}[htb]
\centering
\caption{\textbf{Classification accuracies on SHD using a clean split}, ranked by accuracy.}
\label{tab:shd-perf-table}
\small

\begin{tabular}{p{5cm} c c c c c c}
\toprule
Model & Rec. & Rec. Delays & Ff. Delays & LIF & Param & Test Acc. [\%] \\
\midrule
BRF \citep{higuchi2024balancedresonateandfireneurons} & \cmark &  &  &  & 0.1M & $92.70 \pm 0.70$ \\
SE-adLIF (1L) \citep{Baronig2025} & \cmark &  &  &  & 37.5k & $93.18 \pm 0.74$ \\
EventProp\textit{[b]} \citep{meszaros2025efficienteventbaseddelaylearning} & \cmark & \cmark & \cmark & \cmark & $\sim$1M\textit{[a]} & $93.24 \pm 1.00$ \\
\textbf{DelRec (Only Rec. delays) \textit{Ours}\textit{[b]}} & \cmark & \cmark &  & \cmark & \textbf{0.17M} & \textbf{$93.39 \pm 0.45$} \\
DCLS\textit{[b]} \citep{Hammouamri2024} &  &  & \cmark & \cmark & 0.22M & $93.77 \pm 0.68$ \\
SE-adLIF (2L) \citep{Baronig2025} & \cmark &  &  &  & 0.45M & $93.79 \pm 0.76$ \\
\bottomrule
\end{tabular}

\vspace{2mm}
\footnotesize
\textit{[a]} Estimated from Figure 5.\\
\textit{[b]} Models trained using augmentations.

\end{table}

\subsection{Recurrent delays mitigate the risk of exploding gradients} \label{mitigate explo}

The purpose of this section is to provide a more detailed explanation, without resorting to full mathematical formalism, of why introducing delays in recurrent connections reduces the risk of exploding gradients in these connections. 

Let's consider the generic equation of a neuron with a simple recurrent connection to itself, in the case of a soft reset. We allow the recurrent connection to be delayed by $d$ time-steps, with $d\geq0$:
\begin{equation}
    H[t+1] = \alpha H[t] + I[t+1] + W_{rec}S[t-d] - \theta S[t] \label{eq:base},
\end{equation}
with $S[t] = \Theta(H[t]-\theta)$. Denoting $\mathcal{L}$ the loss function, we can write:
\begin{equation}
    \frac{\partial \mathcal{L}}{\partial W_{rec}} = \sum_{t=0}^T \frac{\partial \mathcal{L}}{\partial H[t]} \cdot \frac{\partial H[t]}{\partial W_{rec}} = \sum_{t=0}^T \delta[t] \cdot \epsilon[t], \label{eq:grad}
\end{equation}
with $\delta_t = \frac{\partial \mathcal{L}}{\partial H[t]}$ and $\epsilon_t = \frac{\partial H[t]}{\partial W_{rec}}$.
\\

We focus on the $\epsilon$ terms. Using Eq.~\ref{eq:base}, we derive:
\begin{equation}
    \epsilon_{t+1} = \alpha \epsilon_t + S[t-d] + W_{rec}\frac{\partial S[t-d]}{\partial W_{rec}} - \theta\frac{\partial S[t]}{\partial W_{rec}},
\end{equation}
Then, noticing that:
\begin{equation}
    \frac{\partial S[t]}{\partial W_{rec}} = \frac{\partial S[t]}{\partial H[t]} \frac{\partial H[t]}{\partial W_{rec}} = \psi'(H[t]-\theta) \epsilon_t,
\end{equation}
with $\psi$ our surrogate function, the relation can be rewritten:
\begin{equation}
    \epsilon_{t+1} = \alpha \epsilon_t + S[t-d] + W_{rec}\psi'(H[t-d]-\theta) \epsilon_{t-d} - \theta \psi'(H[t]-\theta) \epsilon_t.
\end{equation}
Using $\Psi'_{t} = \psi'(H[t]-\theta)$, we find the following recurrence:
\begin{equation}
    \epsilon_{t+1} =  (\alpha - \theta \Psi'_t)\epsilon_t + W_{rec} \Psi'_{t-d} \epsilon_{t-d} + S[t-d].
\end{equation}
At this stage, we simplify the notations by setting $A_t = (\alpha - \theta \Psi'_t)$ and $B_t = W_{rec} \Psi'_{t}$ , to get the simple recurrent relation:
\begin{equation}
    \epsilon_{t+1} = A_t \epsilon_t + B_{t-d} \epsilon_{t-d} + S[t-d].
\end{equation}
One can notice that the range of values taken by $A_t$ and $B_t$ is finite, so we can set:
\begin{align}
    a & = \sup_t |A_t|  , \; b = \sup_t |B_t|,
\end{align}
and have:
\begin{equation}
    |\epsilon_{t+1}| \leq a |\epsilon_t| + b |\epsilon_{t-d}|.
\end{equation}
We admit here that the asymptotic behavior of $\{\epsilon_t\}$ depends solely on the homogeneous term $A_t \epsilon_t + B_{t-d} \epsilon_{t-d}$, so it suffices to focus on studying the sequence:
\begin{equation}
    a\epsilon_t + b\epsilon_{t-d} = \epsilon_{t+1} \label{eq:inf_ineq},
\end{equation}
with positive initial conditions in order to get the asymptotic behavior of $\{\epsilon_t\}$. Thus we define the positive sequence $\{e_t\}$ with the same initial conditions as $\{\epsilon_t\}$, ie $e_0 = |\epsilon_0|,...,e_d=|\epsilon_d|$, and such that:
\begin{equation}
    e_{t+1} = ae_t + be_{t-d}. \label{eq:e_rec}
\end{equation} 
Its characteristic equation is:
\begin{align}
    \lambda^{t+1} = a \lambda^t + b \lambda^{t-d}
    \Longleftrightarrow \lambda^{d+1} - a \lambda^d -b = 0 \Longleftrightarrow \lambda^d(\lambda-a) = b. \label{eq:carac}
\end{align}
Let $\lambda_1,\dots,\lambda_{d+1}$ be the roots of Eq.\ref{eq:carac}, counted with multiplicities. We know that, if all roots are simple, the general solution of Eq.\ref{eq:e_rec} is of the form:
\begin{equation}
    e_{t} = \sum_{k=1}^{d+1} c_k \lambda_k^t,
\end{equation}
and if $\lambda$ is a root of multiplicity $m > 1$, then its contribution to $e_t$ is:
\begin{equation}
    (c_0 + c_1 t+ \dots \ c_{m-1}t^{m-1})\lambda^t.
\end{equation}
Therefore, the asymptotic behavior of $\{e\}$ depends on $r_d = \max_j |\lambda_j|$, which is the spectral radius of the characteristic equation.
\\

If all roots satisfy $|\lambda_j|<1$, then naturally $e_t \longrightarrow 0$. We can also show that it suffices for our gradients ${\epsilon_t}$ to vanish, but we focus here on the gradient explosion. Now, if we are in the case where at least one $|\lambda_j| >1$, then the sequence $\{e_t\}$ is threatened by explosion. In that case, and admitting that $r_d$ is a real simple root of Eq.\ref{eq:carac}, we have:
\begin{equation}
    r_d^d(r_d-a)=b. \label{eq:final_relation}
\end{equation}
Here, one can notice that as $d$ increases, $r_d$ must decrease for Eq.\ref{eq:final_relation} to hold. In other words, increasing the delay $d$ decreases the spectral radius, which slows down the explosion of the solutions of Eq.\ref{eq:carac}. By that mean, increasing $d$ also reduces the growth of the sequence $\{e_t\}$ and thus slows down the explosion of ${\epsilon_t}$.

\putbib

\end{bibunit}

\end{document}